\documentclass[lettersize,journal]{IEEEtran}
\usepackage{amsmath,amsfonts}
\usepackage{algorithmic}
\usepackage{algorithm}
\usepackage{array}
\usepackage[caption=false,font=normalsize,labelfont=sf,textfont=sf]{subfig}
\usepackage{textcomp}
\usepackage{stfloats}
\usepackage{url}
\usepackage{verbatim}
\usepackage{graphicx}
\usepackage{longtable}
\usepackage{cite}
\usepackage{hyperref}
\usepackage{multirow}
\usepackage{booktabs}
\usepackage{caption}
\begin{document}

\title{\LARGE{Integrating YOLO11 and Convolution Block Attention Module for Multi-Season Segmentation of Tree Trunks and Branches in Commercial Apple Orchards}}

\author{%
Ranjan Sapkota,
Manoj Karkee\thanks{$^{*}$Center for Precision and Automated Agricultural Systems, Department of Biological Systems Engineering, Washington State University, USA. Corresponding authors: \texttt{ranjan.sapkota@wsu.edu}, \texttt{manoj.karkee@wsu.edu}}
}

\maketitle

\begin{abstract}
\begin{figure}[!t]
\centering
\includegraphics[width=0.99\columnwidth]{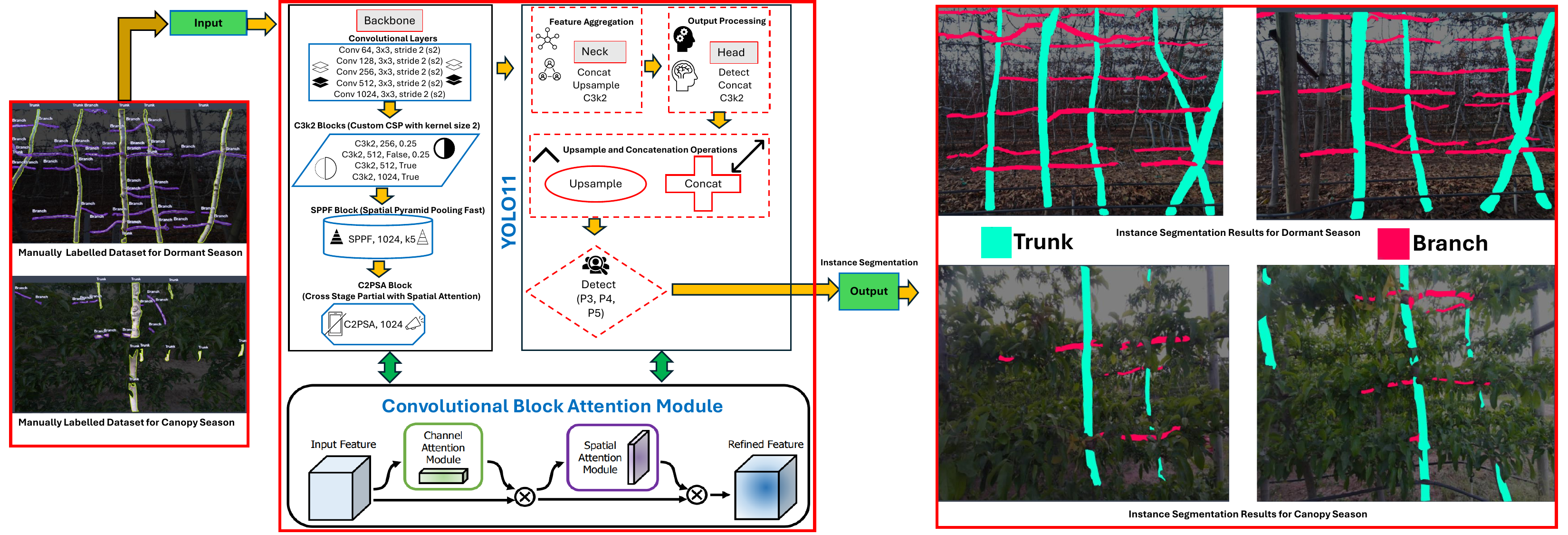} 
\captionof{figure}{Visualization of the YOLO11 and CBAM integration-based instance segmentation approach of apple tree trunks and branches in multiple seasons}
\label{fig:abstractimg}
\end{figure}
In this study, we developed a customized instance segmentation model by integrating the Convolutional Block Attention Module (CBAM) with the YOLO11 architecture. This model, trained on a mixed dataset of dormant and canopy season apple orchard images, aimed to enhance the segmentation of tree trunks and branches under varying seasonal conditions throughout the year. The model was individually validated across dormant and canopy season images after training the YOLO11-CBAM on the mixed dataset collected over the two seasons. Additional testing of the model during pre-bloom, flower bloom, fruit thinning, and harvest season was performed.  The highest recall and precision metrics were observed in the YOLO11x-seg-CBAM and YOLO11m-seg-CBAM  respectively. Particularly, YOLO11m-seg with CBAM showed the highest precision of 0.83 as performed for the Trunk class in training, while without the CBAM, YOLO11m-seg achieved 0.80 precision score for the Trunk class. Likewise, for branch class, YOLO11m-seg with CBAM achieved the highest precision score value of 0.75 while without the CBAM, the YOLO11m-seg achieved a precision of 0.73.  For dormant season validation, YOLO11x-seg exhibited the highest precision at 0.91. Canopy season validation highlighted YOLO11s-seg’s superior precision across all classes, achieving 0.516 for Branch, and 0.64 for Trunk. The modeling approach, trained on two season datasets as dormant and canopy season images, demonstrated the potential of the YOLO11-CBAM integration to effectively detect and segment tree trunks and branches year-round across all seasonal variations. 
\end{abstract}



\section{Introduction}
Labor challenges have significantly impacted the agricultural sector, becoming a critical concern \cite{bochtis2020agricultural, gregorioa2020assessing, calvin2010us}. In 2022, labor costs in this sector reached \$42.57 billion, accounting for nearly 10\% of total production expenses according to a report by the U.S. Department of Agriculture's Economic Research Service \cite{usdaSTATS}. Specifically, the U.S. apple orchards are facing a severe labor shortage, a situation that has deteriorated since the onset of the COVID-19 pandemic \cite{larue2020labor, keegan2023essential}. 

\begin{figure*}[ht]
\centering
\includegraphics[width=0.82\linewidth]{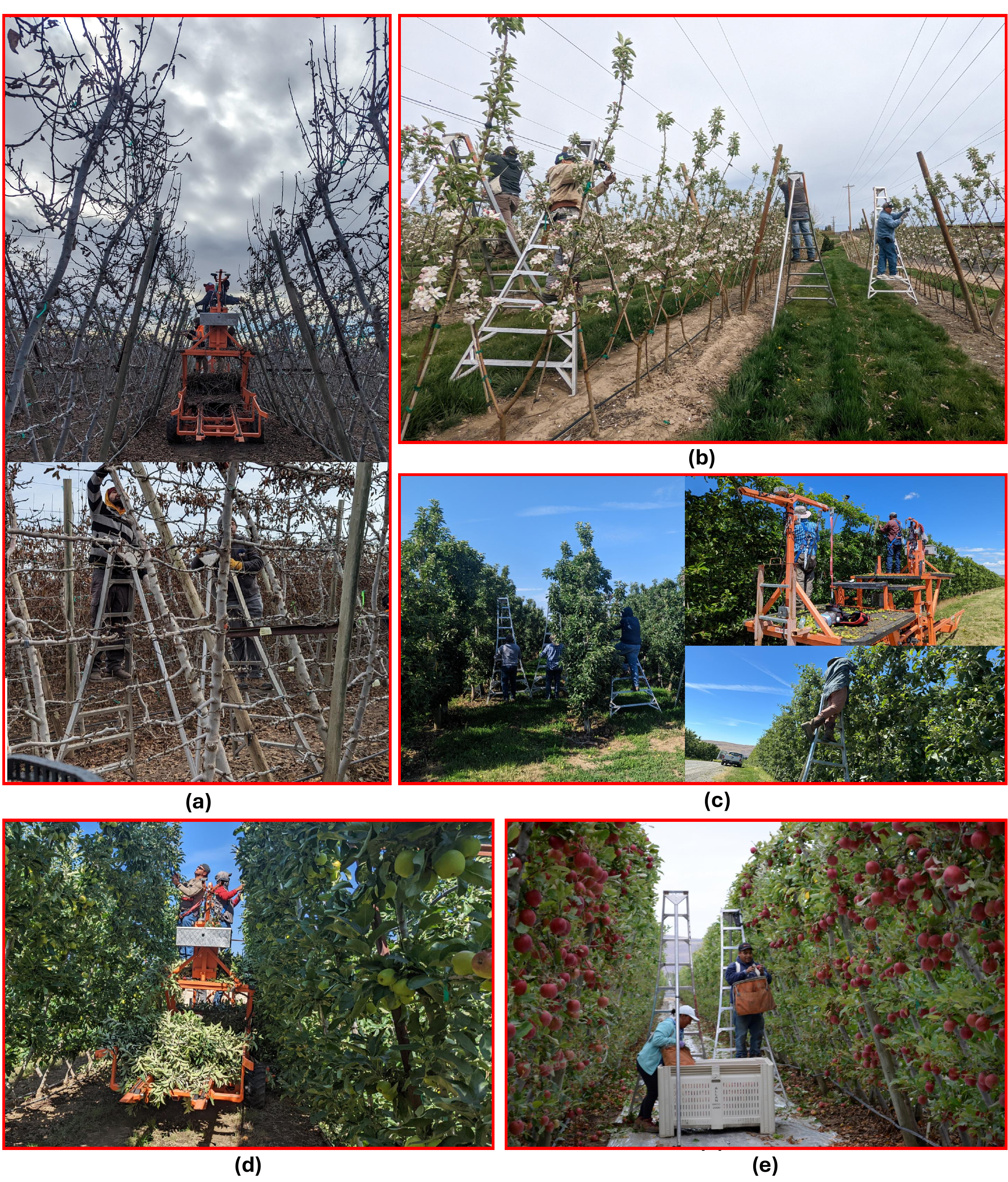}
\caption{) Displays dormant season (December/January), showcasing essential winter pruning for optimal tree health and structure. b) Highlights blossom season (March/April), with manual training of shoots on trellis wires for commercial tree architecture. c) Depicts the thinning of excess green fruits in June/July, critical for quality fruit development. d) Shows summer pruning in August to enhance canopy airflow and light distribution. e) Illustrates the labor-intensive apple harvest in October.}
\label{fig:ProblemStatement}
\end{figure*}

Currently, most of the crop load management operations in commercial apple orchards are manually performed. For example, as observed in a commercial orchard in Prosser, Washington, USA,  Figure \ref{fig:ProblemStatement} a depicts the dormant season (December/January) in a commercial apple orchard, where the trees are leafless, and branches and trunks are fully exposed. The dormant season pruning is vital for several reasons. First, it enhances wound healing and minimizes disease transmission risks, as pathogens are less active during this period \cite{turechek2004apple, marini2018apple}. It also helps shape the tree for optimal structure \cite{wagenmakers1988effects, matias2023citrus} and promotes the growth of new, fruit-bearing branches by directing the tree’s stored energy to selected limbs \cite{mika2011physiological}.

Likewise, Figure \ref{fig:ProblemStatement} b shows the orchard during blossom season (March/April), showing farmers training new shoots along trellis wires to maintain desirable tree architecture for commercial production. Training apple tree shoots to attach to a trellis wire architecture is crucial for optimizing fruit production, improving light penetration, and maintaining a desired tree shape \cite{robinson1998v, nasrabadi2022regulations}. This process encourages the formation of flower buds, allows sunlight to reach more of the canopy, and provides structural strength to the tree \cite{dhillon2014canopy}. The practice requires heavy labor because it involves carefully selecting, positioning, and tying individual shoots to the trellis wires, as well as regular pruning and maintenance to maintain the desired tree form \cite{musacchi2021training, majeed2020deep}.

Similarly, Figure \ref{fig:ProblemStatement} c, shows another crop load management operation where intensive labor is dedicated to thinning green fruits that exceed the tree’s capacity. This operation is critical as it ensures remaining fruits grow to optimal size and quality by reducing competition for resources \cite{link2000significance, sapkota2024immature}. This process helps minimize biennial bearing, promoting consistent annual cropping and improving orchard profitability over time \cite{campbell2024strategies, kumar2021alternate}. Additionally, effective thinning enhances return bloom, which is essential for maintaining productive apple trees in subsequent seasons \cite{costa2018fruit}. 

Likewise, Figure \ref{fig:ProblemStatement} d shows an additional pruning phase in August after fruit thinning, where the orchard is covered by heavy density of tree canopy foliage. This pruning, also known as summer pruning \cite{marini1987summer},  is indeed primarily done to improve air flow and sunlight penetration through the canopy \cite{mupambi2018influence}. By thinning out dense foliage, summer pruning allows better light distribution and maximizes photosynthesis in shaded areas of the tree \cite{dhillon2014canopy, li2001physiological}. Additionally, it reduces branch crowding, which enhances air circulation, decreases humidity, and improves coverage of protectant products like pesticides \cite{teskey2012tree, marini1987summer}. Summer pruning also helps control tree vigor, as it is a devigorating process that reduces carbohydrate production by removing photosynthetic leaf area \cite{nasrabadi2022regulations}. 

Furthermore, Figure \ref{fig:ProblemStatement} e  shows the harvest season when ripe apples are picked, another labor-intensive period requiring significant manpower. Hand picking is the primary method for harvesting fresh market apples, requiring workers to carefully select and remove ripe fruits from trees. This process demands physical effort, dexterity, and the ability to assess fruit quality on the spot. In traditional orchards, pickers often use ladders to reach apples on taller trees, adding to the labor intensity and introducing safety concerns. 

Although these various crop load management operations depicted in \ref{fig:ProblemStatement} are essential for maintaining orchard health and optimizing fruit production, these labor-intensive tasks pose significant health risks, including spinal injuries from repetitive motions \cite{may2020occupational, barneo2021musculoskeletal, ouattara2023agricultural}. The demanding nature of these jobs, combined with the isolation of many orchards, has led to a decline in interest among younger workers, who prefer urban employment opportunities \cite{yeboah2020hard, wu2019future}. Seasonal work exacerbates labor shortages, providing only temporary employment that fails to attract a consistent workforce \cite{alford2017multi, christiaensen2021future, cingolani2024agricultural}.

Moreover, the transient nature of these roles forces many agricultural workers to seek additional income sources, contributing to ongoing labor shortages that challenge orchard operations and impact both the quality and quantity of the harvest \cite{arcury2020latinx, van2022washington, fulcher2023overcoming}. The labor crisis in U.S. agriculture has intensified, particularly after the COVID-19 pandemic, as orchard growers struggle to find enough laborers for essential tasks, risking substantial economic losses and diminished food production \cite{tougeron2021impact, ridley2021effects, larue2020labor, keegan2023essential, sapkota2024yolov10}. This situation poses not only operational challenges but also broader implications for global food security, as the increasing global population heightens the demand for efficient and sustainable agricultural practices \cite{sadigov2022rapid, ben2021artificial}.

The escalating labor shortage in agriculture has highlighted the urgent need for automation technologies, propelling the adoption of robotics to address complex crop load management tasks, particularly in challenging environments like apple orchards. The integration of advanced instance segmentation techniques into robotic systems is proving to be a viable solution to these challenges, offering a revolutionary approach to agricultural practices as demonstrated by Zhang et al. \cite{zhang2024automated}. Visual perception and understanding are crucial in these settings, where robots must navigate and operate precisely. Key to this capability is the effective detection and segmentation of tree trunks and branches, as outlined by Majeed et al. \cite{majeed2018apple} and Kok et al. \cite{kok2023obscured}, which are essential for enabling robots to automate critical tasks such as pruning \cite{zahid2021technological} and fruit picking \cite{li2020detection}. Instance segmentation, a critical technology within computer vision, merges object detection with semantic segmentation to deliver detailed, pixel-level insights, as discussed by Hafiz \cite{hafiz2020survey}. This technology is pivotal in agriculture for advancing automated and robotic operations, enhancing productivity and sustainability. By providing a detailed and precise localization of plant structures, instance segmentation facilitates comprehensive analyses of plant growth and health, thereby influencing everything from yield estimation to disease management \cite{hafiz2020survey}. Such capabilities are indispensable for developing precise, targeted interventions that can significantly improve pest management and the efficient use of water and nutrients \cite{zhang2020applications, zhang2020efficient}, optimizing resource use and minimizing waste.

The detection and segmentation of tree trunks and branches are crucial in the effective management of crop loads in orchards, as depicted in various operations such as green fruit thinning and flower thinning, shown in Figure \ref{fig:ProblemStatement}. The accurate identification of these structural elements is vital for executing precision tasks by automated systems. For instance, in the context of pruning, as illustrated in Figure \ref{fig:ProblemStatement}a, the measurement of branch diameter—which correlates to limb cross-sectional area (LCSA)—is essential. Research suggests that maintaining about 5 fruits per cm\textsuperscript{2} of LCSA optimizes crop yield and tree health \cite{sidhu2022crop}, although practices vary, with some farmers opting for 6 fruits per cm\textsuperscript{2} LCSA \cite{scalisi2024localised, voglhuber2022setting}. The precise segmentation of tree trunks and branches helps make decisions about pruning and thinning, focusing on evenly distributing fruits to improve both the quality and quantity of the yield. Accurate branch measurements support computer models that manage these automated tasks, underscoring the role of machine vision in current farming techniques.

Furthermore, the localization of tree branches is critical for the operation of robotic arms and machinery within the complex spatial layouts of orchards, ensuring efficiency and safety in mechanical operations \cite{chen2018multi, brown2024tree, shalal2015orchard}. For example, during summer pruning and harvesting operations, as shown in Figures \ref{fig:ProblemStatement}d and \ref{fig:ProblemStatement}e, knowing the exact positions of trunks and branches allows for precise cuts and effective fruit picking by robots, minimizing physical contact that could damage the trees or fruits.

In this study, we introduce a novel approach that diverges from traditional seasonal modeling of tree trunk and branch segmentation. Our approach challenges the conventional methods by proposing a continuous, season-independent model. Instead of modeling the vision system individually for dormant and canopy season images, we propose merging images from two distinct seasonal phases, complete dormancy (in December) and high-density canopy (in June), into a single dataset. This approach involves labeling all visible trunk and branch structures within the frontal section of the orchard rows, categorizing them consistently as either trunk or branch across all images. By integrating these diverse seasonal representations into a unified model, we aim to develop a robust machine vision system capable of effective segmentation throughout varying seasonal conditions. Our hypothesis posits that a model trained on images captured during both fully dormant and high-density canopy foliage seasons should be effective across all seasonal variations. This approach is grounded in the belief that robust training on these extreme states will equip the model to handle intermediate stages like partial canopy, flower bloom, or harvesting season throughout the year. 

The specific objectives of this study are:
\begin{itemize}
\item To collect images during the dormant and canopy seasons using a consumer-grade machine vision camera mounted on a robotic platform. The platform will navigate a commercial apple orchard with manual intervention to ensure comprehensive coverage.
    \item To annotate tree trunks and branches manually, integrating images from both seasons into a unified dataset folder to create a robust class for research purposes.
    \item To integrate the Convolutional Block Attention Module (CBAM) with all five configurations of the YOLO11 object detection and segmentation model. This integration aims to enhance the accuracy of detecting and segmenting tree trunks and branches.
    \item To validate the model trained on the mixed dataset by testing it across the individual seasons of Dormant and Canopy, with a total of 78 images from each season to assess performance consistency and accuracy.
    \item To test the model's performance on tree trunk and branch segmentation across four additional datasets representing different seasonal phases: before flower bloom, during flower bloom, during green fruit thinning, and during the apple harvesting season.
\end{itemize}

\section{Methods}
\begin{figure*}[ht]
\centering
\includegraphics[width=0.95\linewidth]{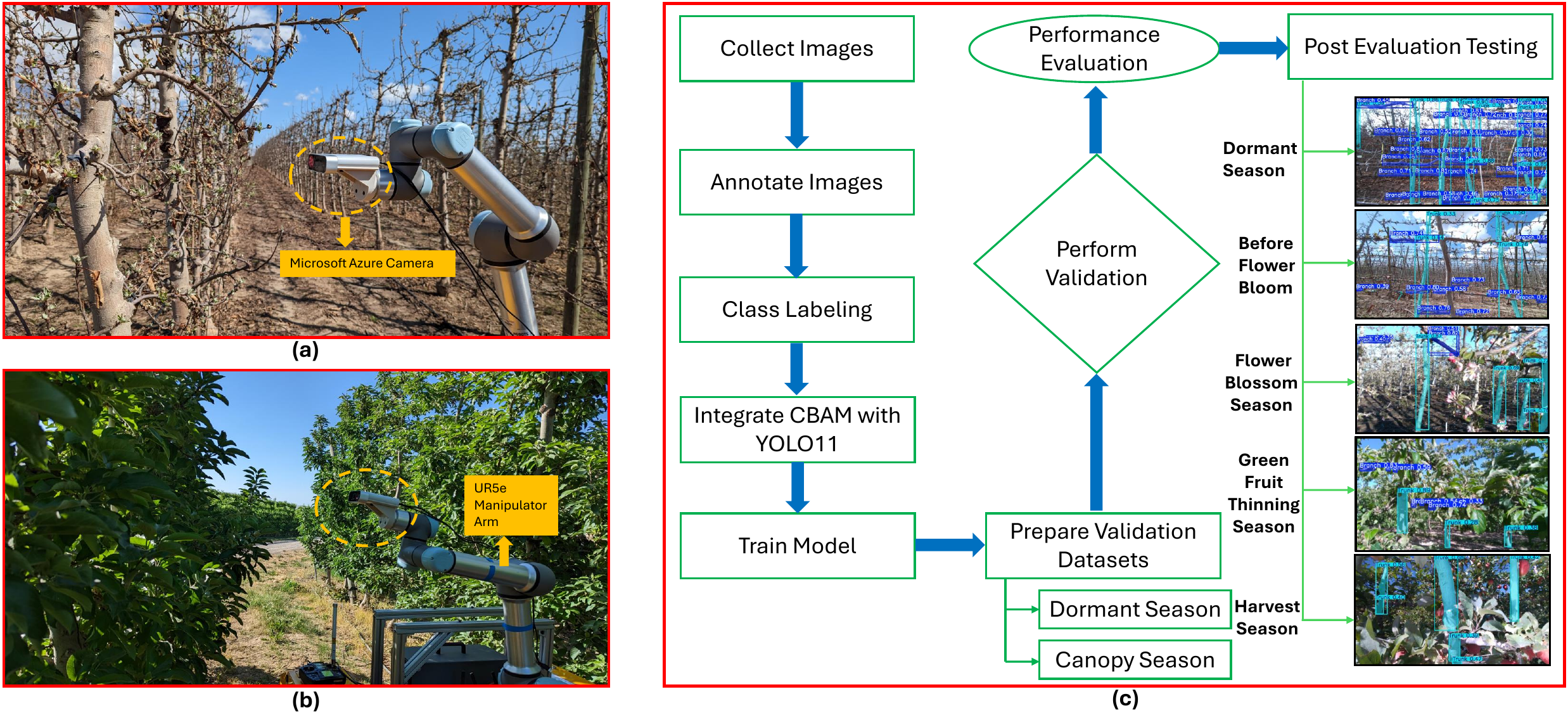}
\caption{Showing the Image Acquisition and Methodology workflow: a) Dormant season image collection using Microsoft Azure machine vision camera ; b) Canopy season image collection using Microsoft Azure machine vision camera ; c) Workflow diagram }
\label{fig:method1}
\end{figure*}

In this study, a unique modeling approach was adopted where images from two distinct seasons—dormant (Figure \ref{fig:method1}a) and canopy seasons (Figure \ref{fig:method1}b) were collectively utilized. These images were annotated to label all visible trunks and branches, irrespective of the season, classifying them into trunk and branch classes. Subsequently, the CBAM was integrated with five configurations of the YOLO11 model to train the system for instance segmentation across the two seasonal backgrounds. Post-training, each of the five YOLO11 configurations, augmented with CBAM, was validated using two separate datasets as shown in (Figure \ref{fig:method1}c) These datasets comprised 78 images from the dormant season and 78 images from the canopy season. Each image had been previously labeled, and these labels served as the basis for validation. The model was first validated using a dataset that combined images from two distinct seasons. It was then tested across various stages of the growing cycle, starting with the pre-blossom season, followed by the flower blossom season, progressing through the green fruit thinning season, and finally the harvest season.
\subsection{Study Site and Data Acquisition}
This study was conducted in a commercial apple orchard owned and operated by Allan Brothers Fruit Company, located at Prosser, Washington State, USA. The orchard was planted in 2009 with a Scilate apple cultivar with a row spacing of 9.0 ft., and a plant spacing of 3.0 f. and was trained to a V-trellis architecture. Images were collected using a Microsoft Azure camera, which was mounted over a UR5e Manipulator Arm as shown in Figure \ref{fig:method1}. In January 2023, dormant season images were captured, as depicted in Figure \ref{fig:method1}a. Subsequently, in June 2024, canopy season images were acquired as shown in Figure \ref{fig:method1}b. 

\subsection{Data Preparation for Model Training}
A comprehensive set of 859 images was amassed for the study, consisting of 553 images from the canopy season and the remainder from the dormant season. All images were consolidated into a single folder to streamline the data preparation process. To enhance the dataset and prepare it for effective model training, a series of image augmentations were applied. This augmentation not only diversified the visual data but also simulated various photographic conditions to robustly train the segmentation model. The applied augmentations resulted in each training example being outputted thrice, effectively increasing the number of training images. Specific transformations included a 90° rotation, which inverted the images, introducing a new orientation challenge. Horizontal shearing was kept at 0° to maintain alignment, while vertical shearing was varied by ±15° to simulate different angles of viewing the apple trees. The hue of the images was adjusted between -15° and +15°, altering the color perception slightly to mimic different lighting conditions. 

Additionally, saturation adjustments ranged from -25\% to +25\%, modifying the color intensity to reflect varying weather conditions. Brightness and exposure were both altered between -20\% and +20\%, replicating different times of the day and cloud cover scenarios. These augmentations expanded the initial dataset to a total of 2070 images designated for training, with 84 images set aside for validation and 85 for testing.
\begin{figure*}[ht]
\centering
\includegraphics[width=0.95\linewidth]{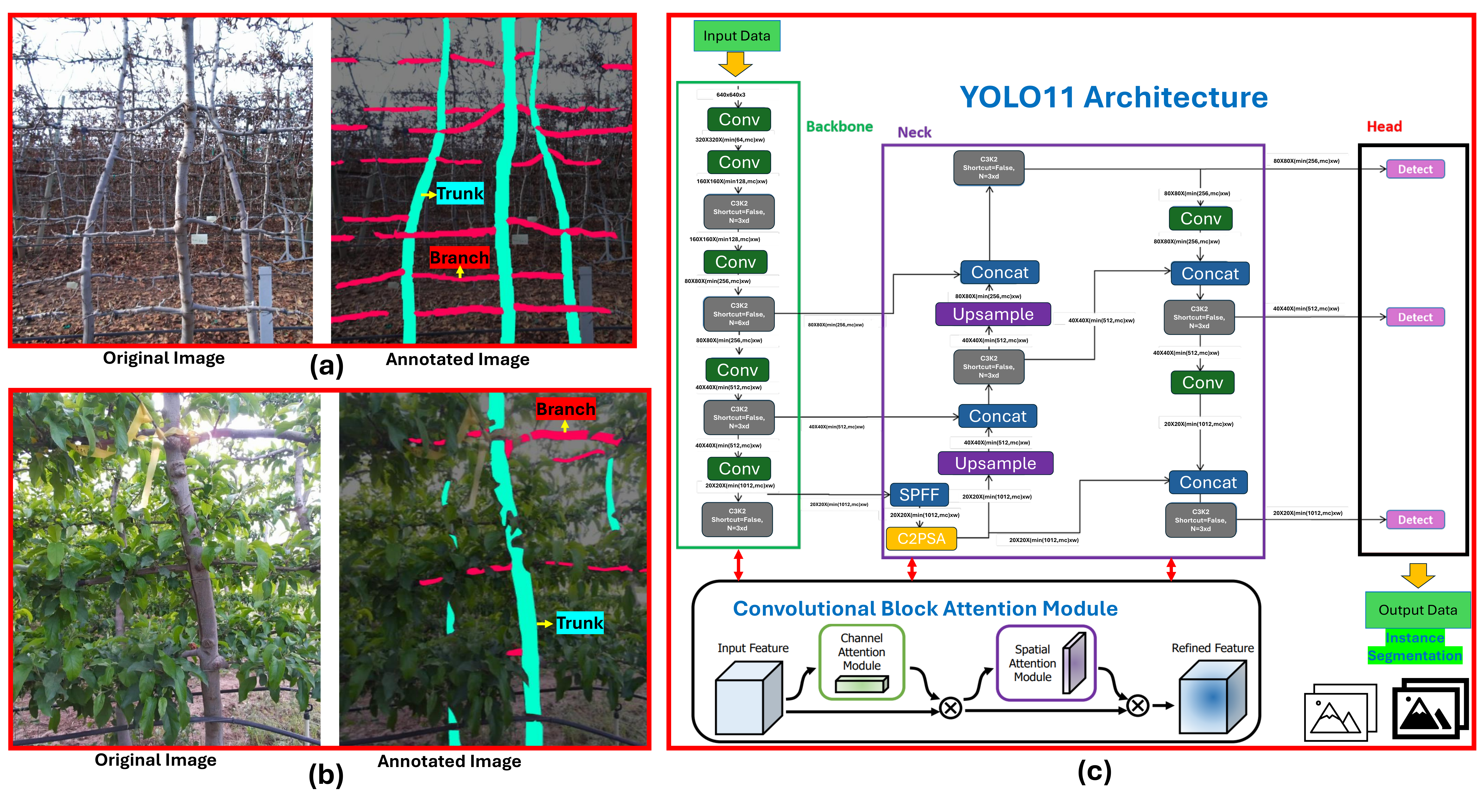}
\caption{Showing the Image Labelling a) Dormant season image labelling into trunks and branch; b) Canopy season image into trunk and branch ; c) Architecture diagram of YOLO11 \cite{sapkota2024comparing, sapkota2024comprehensive, sapkota2024yolo11} fusion implemented in this study }
\label{fig:methoddatatrain}
\end{figure*}
\subsection{Integrate and Train CBAM with YOLO11 for Deep Learning}
Overview of the two models that were fused are as follows:
\subsubsection{CBAM}
The CBAM is designed to enhance the feature sensitivity of convolutional neural networks (CNNs) by focusing on the most informative parts of an image. CBAM operates through two core components: the Channel Attention Module (CAM) and the Spatial Attention Module (SAM) \cite{woo2018cbam}. The CAM determines the significance of each channel in the feature map by leveraging global average and max pooling, processed through a multi-layer perceptron to produce a channel-specific attention map. This helps the model to emphasize the more informative features. Following CAM, the SAM identifies crucial spatial areas by applying pooling operations across channels, enhancing the model's focus on relevant spatial regions. These attention mechanisms refine feature maps through element-wise multiplication, optimizing the network’s performance across various tasks by enabling selective emphasis on important features. This selective focus not only boosts performance but also enhances the interpretability of the CNNs, making CBAM a versatile and effective addition to any CNN architecture \cite{du2022broodstock}.
\subsubsection{YOLO11}
YOLO11 represents the cutting-edge development within the YOLO series, renowned for its efficient and powerful performance in real-time computer vision tasks \cite{sapkota2024zero}. Building upon the foundational strengths of previous iterations, YOLO11 introduces enhancements in feature extraction, significantly improving the detection and segmentation of objects across varied scales and orientations \cite{sapkota2024comparing, sapkota2024comprehensive, sapkota2024yolo11}. This model boasts an optimized architecture that ensures efficient processing, enabling it to handle complex tasks such as instance segmentation and pose estimation effectively. Notable for its high mean Average Precision (mAP) score and reduced parameter count compared to its predecessors, YOLO11 provides substantial accuracy improvements while maintaining faster processing speeds. This makes YOLO11 particularly suitable for applications requiring real-time analysis, such as surveillance and agricultural monitoring, where timely and accurate object recognition is crucial. The enhancements in YOLO11 underline its adaptability and potency in tackling a broad array of vision-based challenges.

\subsubsection{YOLO11-CBAM fusion}
The integration began with the definition of two key components within the CBAM: the Channel Attention Module (CAM) and the Spatial Attention Module (SAM). CAM was designed to prioritize channel-wise features by utilizing global average pooling and global max pooling. These pooled outputs were then processed via a multi-layer perceptron, producing a channel attention map that signals which channels contain more relevant information. This process helps the model focus on features that are crucial for identifying tree structures more accurately. Following the CAM, the SAM was employed to pinpoint critical spatial regions within the images. It achieved this by applying average and max pooling across the channel dimension, followed by a concatenation of these results. A convolutional layer then processed these combined features to produce a spatial attention map. This map guides the model to focus on significant spatial details, enhancing its ability to segment tree trunks and branches effectively. 

The YOLO11 model, particularly the YOLO11n configuration, was then modified to incorporate these CBAM enhancements. The integration involved embedding the CBAM modules right after each convolutional layer within the YOLO11 architecture. This structural modification was systematically applied across all other YOLO11 configurations (s, m, l, x) to maintain consistency in performance enhancements. Training parameters were set to optimize the learning process, with the training executed over 500 epochs on a dataset comprising a mixed collection of dormant and canopy season images. This dataset was annotated and formatted specifically for YOLO11, ensuring that the model was trained on diverse seasonal representations. The performance of each configuration was then meticulously evaluated across a series of metrics: precision, recall, mean Average Precision (mAP@50), alongside processing speeds (pre-processing, post-processing, and inference). 
\subsection{Performance Metrics Evaluation}

During the training stage, the performance of each model configuration was thoroughly assessed in terms of precision, recall, and mAP@50. These metrics were derived from the following fundamental equations:

\begin{equation}
    \text{Precision} = \frac{TP}{TP + FP}
\end{equation}

\begin{equation}
    \text{Recall} = \frac{TP}{TP + FN}
\end{equation}

\begin{equation}
    \text{MIoU} = \frac{\text{Area of Overlap}}{\text{Area of Union}} = \frac{TP}{FP + TP + FN}
\end{equation}

Where:
\begin{itemize}
    \item \( TP \) (True Positives) denotes the number of correctly identified objects (trunks and branches) in both dormant and canopy seasons.
    \item \( FP \) (False Positives) indicates the number of incorrect predictions where non-target objects were misidentified as trunks or branches.
    \item \( FN \) (False Negatives) represents the instances where actual trunks or branches present in the images were missed by the model.
\end{itemize}

Precision quantifies the accuracy of the positive predictions, while recall measures the model’s ability to identify all relevant instances. The mAP@50 provides an overall effectiveness of the model at a 50\% Intersection Over Union (IOU) threshold, reflecting how well the model distinguishes between the tree trunk and branch classes across different seasons.

Additionally, the model’s preprocessing, postprocessing, and inference speeds were evaluated to gauge the operational efficiency.
Furthermore, the total number of convolution layers, training time for each configuration, and GFLOPs (Giga Floating Point Operations per Second) were meticulously calculated, offering insights into the computational demands and efficiency during the training process involving mixed seasonal images of tree trunks and branches. GFLOPs provide a measure of the computational load during the inference phase, thereby giving insights into the model's efficiency and processing speed:

\begin{equation}
    GFLOPs = \frac{\text{Total floating-point operations per inference}}{10^9}
\end{equation}

Where:
\begin{itemize}
    \item \textbf{Total floating-point operations per inference}: This is the sum of all operations that involve floating-point calculations performed during a single forward pass of one image through the neural network. In the context of your dataset, this includes operations required for processing images of tree trunks and branches across different seasons, utilizing convolutional layers for feature extraction and attention mechanisms to enhance detection accuracy.
    \item \textbf{\(10^9\)}: This factor converts the total operations into giga (billion) floating-point operations, standardizing the measure to GFLOPs.
\end{itemize}

\subsection{Prepare Validation Dataset and Evaluate the Model's Performance across Each Season}
For the purpose of validation, datasets comprising 78 images each from dormant and canopy seasons were meticulously prepared. These images were manually annotated using the Roboflow online labeling platform, specifically tailored for the YOLO11 format to facilitate precise model validation. Annotations were carefully crafted to ensure accurate representation of tree trunks and branches during both seasonal phases. Once annotated, two distinct folders were organized, segregating the images according to their respective seasons—dormant and canopy. This separation was crucial for assessing the model's performance under different seasonal conditions, providing insights into its adaptability and accuracy throughout the year. The performance of the model was rigorously evaluated for each set of seasonal images. Metrics such as mask precision, mask recall, and mean average precision (mAP@50) were calculated to determine the efficacy of the model in identifying and segmenting tree structures correctly. Additionally, the speeds of preprocessing, postprocessing, and inference were meticulously measured. These evaluations were essential to understand the operational dynamics of the model, examining its efficiency in processing time and its capability to handle real-time applications.
\begin{figure*}[ht]
\centering
\includegraphics[width=0.99\linewidth]{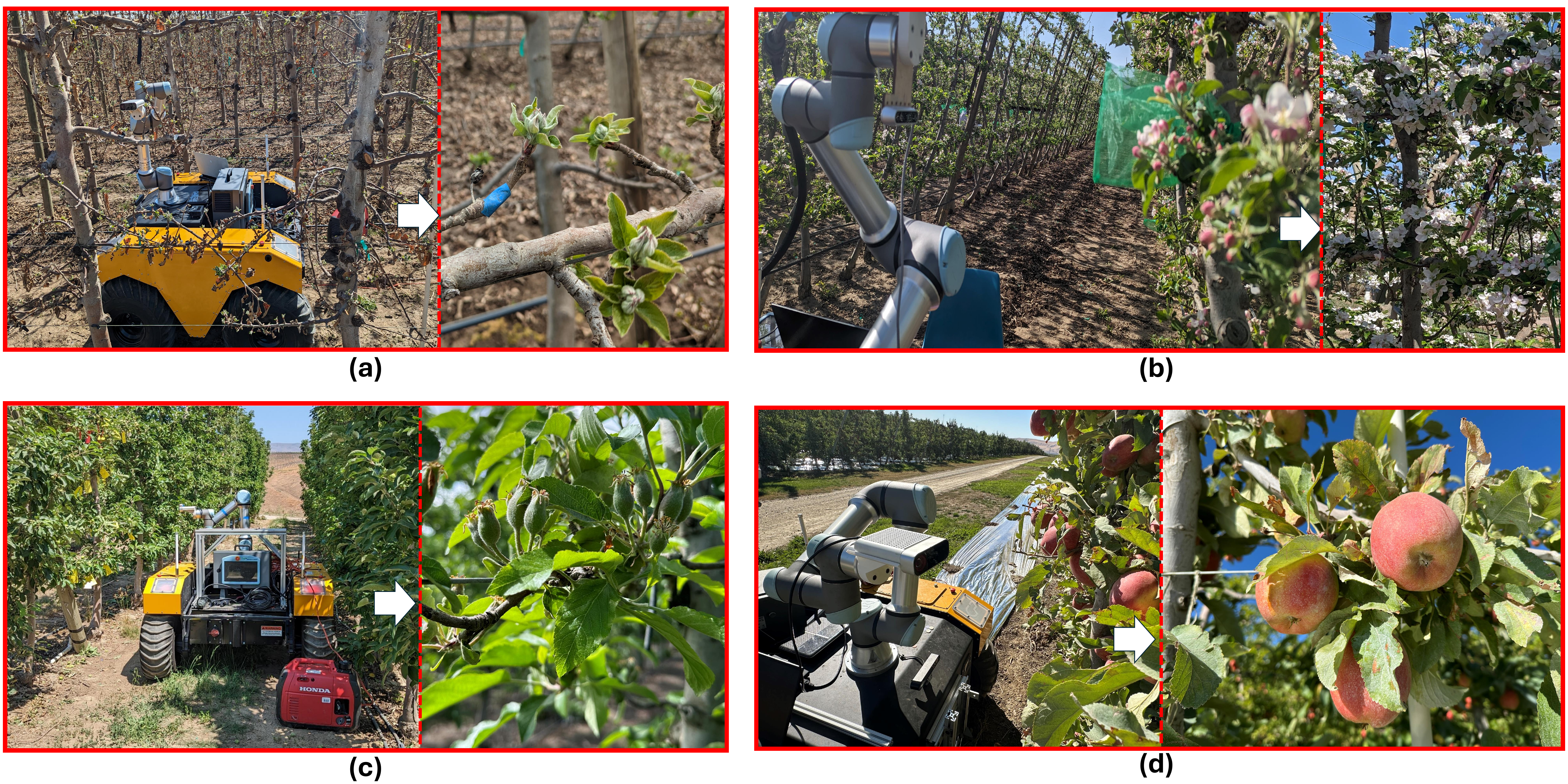}
\caption{Distinct seasonal conditions tested commercial apple orchard: a) Pre-blossom season image collection setup showing early branch development; b) Flower bloom season with active pollination and flower thinning activities; c) Green fruit thinning season highlighting the dense fruitlet clusters needing thinning; d) Harvest season image collection in ripe apples, ready for market.}
\label{fig:tesmethods}
\end{figure*}
\subsection{Testing Across Multiple Seasons}
Following the training and validation of the YOLO11-CBAM model, extensive testing was performed across four distinct seasonal conditions, as illustrated in Figures \ref{fig:tesmethods}a, \ref{fig:tesmethods}b, \ref{fig:tesmethods}c, and \ref{fig:tesmethods}d. These tests were strategically conducted to encompass a range of developmental stages in the apple orchard cycle, including pre-blossom, flower bloom, green fruit thinning, and harvesting seasons. Figure \ref{fig:tesmethods}a captures the early pre-blossom season where the machine vision system was employed to gather visual data in the orchards. The left image displays the setup for image collection during this nascent stage of the orchard cycle, characterized by branches with emerging small leaves and buds yet to bloom. The right image provides a closer view of the branches at this preliminary stage, highlighting the nascent vegetative growth. During the flower bloom season, depicted in Figure \ref{fig:tesmethods}b, further image collection was conducted to document this critical phase of pollination and flower thinning. The left image in Figure \ref{fig:tesmethods}b demonstrates the field setup for capturing this phase, while the right image offers a detailed view of the orchard in full bloom, underscoring the necessity for precise pollination and selective thinning to regulate crop yield.

Likewise, the green fruit thinning season is showcased in Figure \ref{fig:tesmethods}c, where a robotic system equipped with a machine vision system was utilized for image acquisition. The image on the left shows the operational setup in the orchard, while the right image focuses closely on clusters of early-season green fruitlets that require thinning to ensure optimal fruit development and orchard productivity. Furthermore, Figure \ref{fig:tesmethods}d  shows the harvesting season where the left image illustrates the use of the machine vision system in the orchard as ripe apples are ready to be picked for market purpose. The corresponding right image in Figure \ref{fig:tesmethods}d shows a clear view of the ripe apples scene collected using the machine vision camera. 

\section{Results}
\begin{figure*}[ht]
\centering
\includegraphics[width=0.99\linewidth]{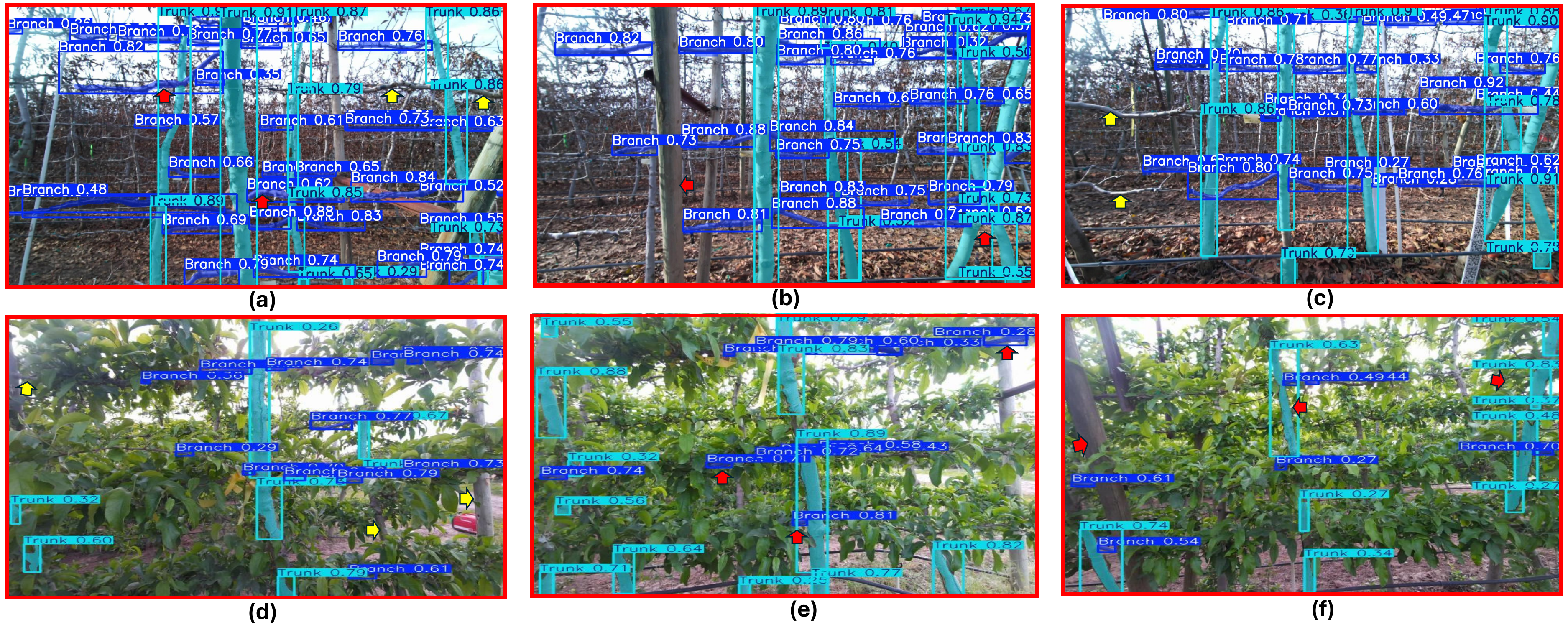}
\caption{Result Examples: (a) Depicts a dormant season scene where the YOLO11 model effectively segments a branch (red arrow) but misses another (yellow arrow). (b) Demonstrates the model's precision in avoiding false trunk identification (red arrow) and highlights missed branches (yellow arrows) in a dormant season context. (c) Shows areas where the model failed to detect prominent branches in the foreground (yellow arrows). (d) Displays a canopy season image, identifying most trunks and branches accurately, though some are missed (yellow arrows). (e) Illustrates successful branch segmentation in dense canopy foliage (red arrows). (f) Highlights accurate trunk segmentation amidst complex canopy coverage (red arrows).}
\label{fig:resultfirst}
\end{figure*}

Figure \ref{fig:resultfirst} presents detection and segmentation outcomes for tree trunks and branches using the YOLO11n-seg algorithm across dormant and canopy seasons within commercial apple orchards. Demonstrating the YOLO11 algorithm's proficiency, the model adeptly delineated trunk and branch structures, utilizing a training set that encompassed images from both seasons. For instance, Figures \ref{fig:resultfirst}a, \ref{fig:resultfirst}b, and \ref{fig:resultfirst}c showcase segmentation results on validation images from the dormant season, applying a training, testing, and validation ratio of 8:1:1. Conversely, Figures \ref{fig:resultfirst}d, \ref{fig:resultfirst}e, and \ref{fig:resultfirst}f illustrate outcomes from the canopy season.

Figure \ref{fig:resultfirst}a features a red arrow indicating a scenario where the YOLO11 model, trained on the combined dataset, accurately identified a branch amid the orchard’s complexity. Conversely, a yellow arrow highlights a missed branch detection despite clear visibility. Figure \ref{fig:resultfirst}b's red arrow on the right illustrates the model's discernment in not mistaking an orchard training pole for a trunk, a common visual confusion. Additionally, on the image's left side, a red arrow marks a successful trunk identification, where the model accurately segmented intersecting trunks without error. In contrast, Figure \ref{fig:resultfirst}c's yellow arrows point to two prominent branches the model failed to detect, despite their apparent foreground position. To enhance accuracy, increasing the number of training examples and ensuring comprehensive labeling of all visible branches and trunks could be beneficial. Omitting no element within the image frame, especially those directly in the camera’s view, may refine detection capabilities. Despite these challenges, the demonstrated outcomes reveal the model's robust potential in adapting to diverse seasonal conditions within orchard environments.

Similarly, Figure \ref{fig:resultfirst}d illustrates a canopy season scenario where the YOLO11 model adeptly detected and segmented the majority of visible trunks and branches. Nonetheless, yellow arrows point out regions where detection and segmentation fell short. Specifically, the top-left yellow arrow indicates a branch that the model overlooked, while two yellow arrows on the right highlight visible trunks that were not identified by the trained model. Continuing this analysis, Figure \ref{fig:resultfirst}e underscores the model’s effectiveness in complex scenarios, marked by dense canopy foliage. Red arrows in this figure exemplify the model’s precision in identifying and segmenting tree branches accurately. Each of the three red arrows in Figure \ref{fig:resultfirst}e represents instances where the YOLO11 successfully segmented branches, demonstrating robust performance in challenging environments typical of the canopy season. Further affirming the model's capabilities, Figure \ref{fig:resultfirst}f features additional instances of precise segmentation. Here, the leftmost red arrow points to a scenario where the model correctly distinguished between a training pole and a tree trunk, avoiding common misidentifications. The middle red arrow highlights successful trunk segmentation, and the rightmost arrow shows the model adeptly segmenting a trunk obscured by dense foliage and canopy.
\subsection{YOLO11-CBAM Training Results}
\subsubsection{\textbf{Performance metrics Evaluation of YOLO11 Instance Segmentation Algorithm during Model Training}}
The YOLO11 instance segmentation models demonstrated varied performance across different configurations and classes. For the "All" class, the highest precision of 0.74 was achieved by YOLO11m-seg, while YOLO11x-seg recorded the highest recall of 0.61. The greatest mAP@50 was scored by YOLO11l-seg, reaching 0.64. In the "Branch" class, YOLO11m-seg outperformed others in mask precision with a score of 0.66, YOLO11x-seg topped mask recall with 0.50, and again, YOLO11m-seg led mAP@50 with a score of 0.52. For "Trunk" segmentation, YOLO11m-seg delivered the highest mask precision at 0.83, YOLO11x-seg achieved the highest recall at 0.72, and YOLO11l-seg topped mAP@50 with 0.77. Detailed performance metrics for precision, recall, and mAP@50 for all five configurations of YOLO11: YOLO11n, YOLO11s, YOLO11m, YOLO11l, and YOLO11x are systematically presented in \ref{tab:performance_metrics} 1, showcasing results across the "All," "Branch," and "Trunk" classes.

\begin{table*}[ht]
\centering
\caption{Showing the performance metrics for YOLO11 instance segmentation models across all, branch, and trunk classes during the training validation phase.}
\label{tab:performance_metrics}
\begin{tabular}{@{}lccccccccc@{}}
\toprule
\textbf{Config} & \textbf{Class} & \multicolumn{4}{c}{\textbf{Box Metrics}} & \multicolumn{4}{c}{\textbf{Mask Metrics}} \\
\cmidrule(lr){3-6} \cmidrule(lr){7-10}
 & & \textbf{P} & \textbf{R} & \textbf{mAP@50} & & \textbf{P} & \textbf{R} & \textbf{mAP@50} & \\
\midrule
\multirow{3}{*}{YOLO11n-seg} & All    & 0.742 & 0.622 & 0.654 & & 0.717 & 0.576 & 0.623 & \\
                             & Branch & 0.693 & 0.533 & 0.563 & & 0.643 & 0.468 & 0.504 & \\
                             & Trunk  & 0.791 & 0.710 & 0.744 & & 0.792 & 0.685 & 0.742 & \\
\midrule
\multirow{3}{*}{YOLO11s-seg} & All    & 0.762 & 0.632 & 0.680 & & 0.717 & 0.589 & 0.630 & \\
                             & Branch & 0.715 & 0.547 & 0.589 & & 0.637 & 0.479 & 0.508 & \\
                             & Trunk  & 0.809 & 0.717 & 0.771 & & 0.796 & 0.699 & 0.753 & \\
\midrule
\multirow{3}{*}{YOLO11m-seg} & All    & 0.766 & 0.661 & 0.698 & & 0.749 & 0.581 & 0.636 & \\
                             & Branch & 0.707 & 0.584 & 0.613 & & 0.669 & 0.475 & 0.516 & \\
                             & Trunk  & 0.825 & 0.738 & 0.783 & & 0.830 & 0.688 & 0.755 & \\
\midrule
\multirow{3}{*}{YOLO11l-seg} & All    & 0.770 & 0.658 & 0.691 & & 0.736 & 0.604 & 0.644 & \\
                             & Branch & 0.720 & 0.576 & 0.600 & & 0.653 & 0.496 & 0.514 & \\
                             & Trunk  & 0.820 & 0.739 & 0.782 & & 0.818 & 0.713 & 0.773 & \\
\midrule
\multirow{3}{*}{YOLO11x-seg} & All    & 0.762 & 0.669 & 0.700 & & 0.700 & 0.614 & 0.641 & \\
                             & Branch & 0.733 & 0.590 & 0.623 & & 0.633 & 0.503 & 0.514 & \\
                             & Trunk  & 0.790 & 0.747 & 0.776 & & 0.768 & 0.724 & 0.769 & \\
\bottomrule
\end{tabular}
\end{table*}

\begin{figure}[ht]
\centering
\includegraphics[width=0.99\linewidth]{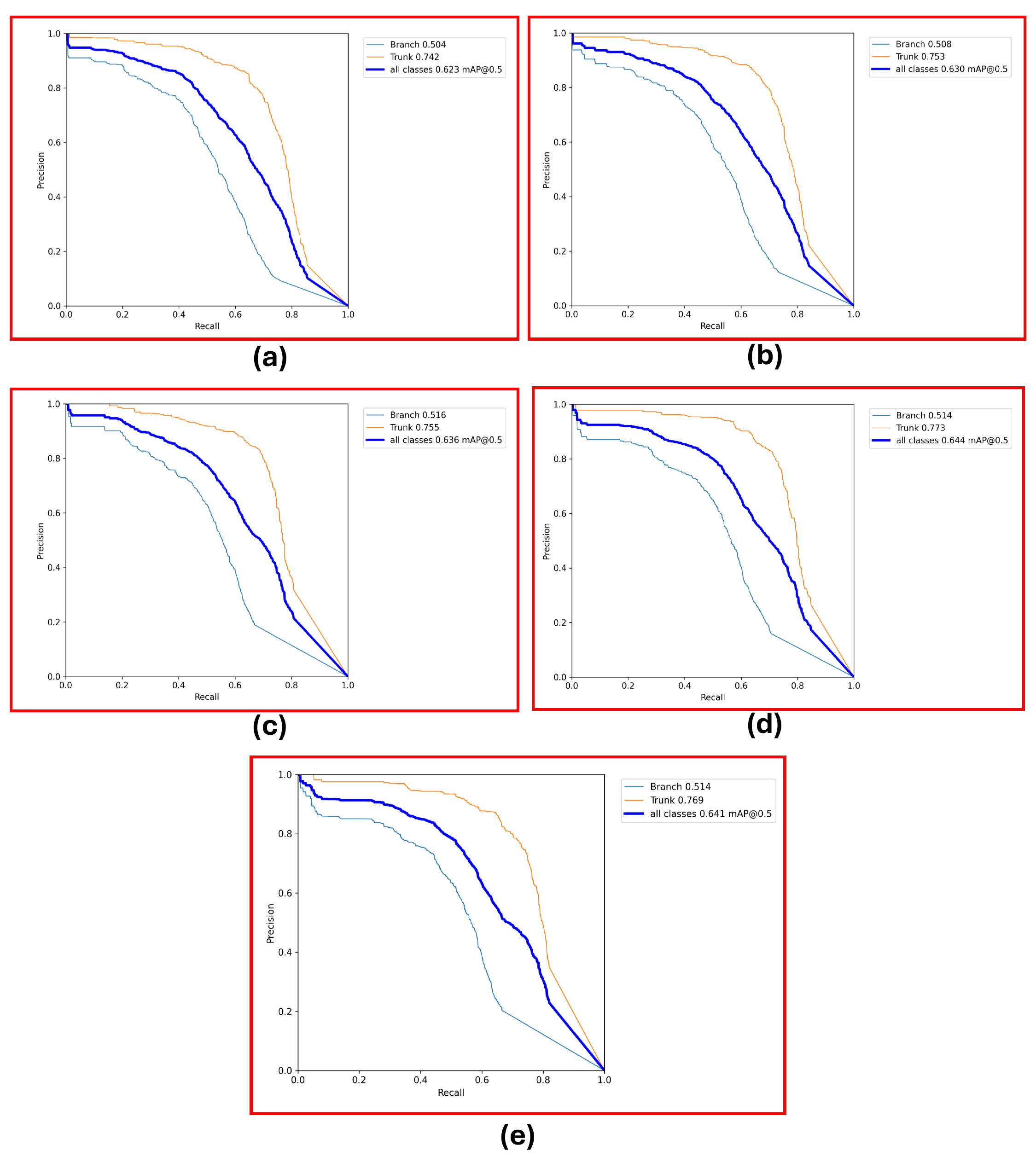}
\caption{Precision Recall-Confidence Curves for Mask Metrics of a) YOLO11n-seg b)YOLO11s-seg; c) YOLO11m-seg ;  d) YOLO11l-seg ; and e) YOLO11x-seg}
\label{fig:PRcurve}
\end{figure}

\begin{figure}[ht]
\centering
\includegraphics[width=0.99\linewidth]{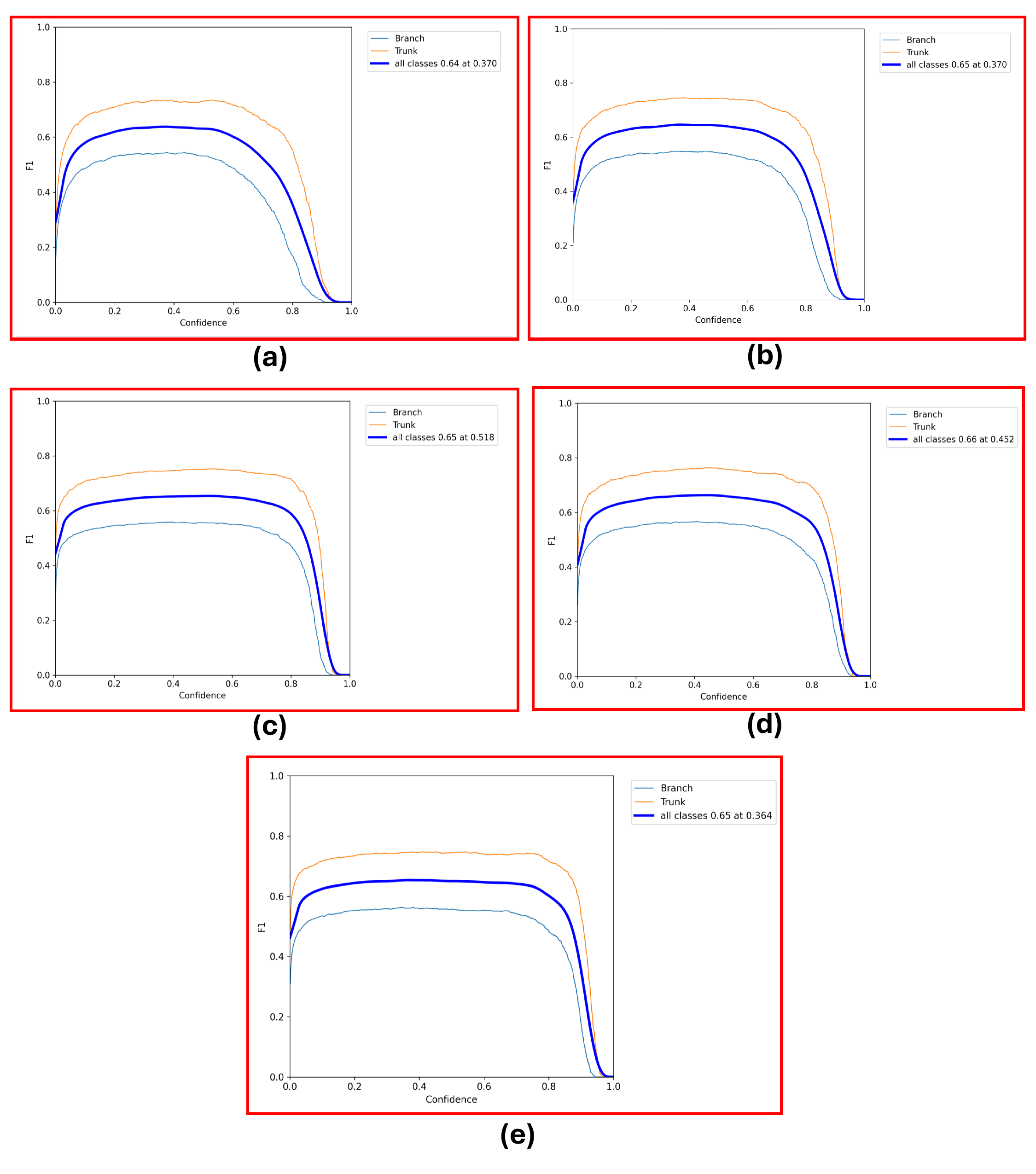}
\caption{F1-confidence curves for a) YOLO11n-seg b)YOLO11s-seg; c) YOLO11m-seg ;  d) YOLO11l-seg ; and e) YOLO11x-seg}
\label{fig:F1curves}
\end{figure}

Figure \ref{fig:PRcurve} displays the precision-recall confidence curves for the five configurations of the YOLO11 segmentation algorithms post-training. Specifically, Figure \ref{fig:PRcurve}a depicts the curve for the YOLO11n-seg configuration, \ref{fig:PRcurve}b for YOLO11s-seg, \ref{fig:PRcurve}c for YOLO11m-seg, \ref{fig:PRcurve}d for YOLO11l-seg, and \ref{fig:PRcurve}e showcases the curve for YOLO11x-seg. Likewise, Figure \ref{fig:F1curves} presents the F1-confidence curves for the five configurations of the YOLO11 segmentation algorithms after training. Specifically, Figure \ref{fig:F1curves}a illustrates the curve for YOLO11n-seg, \ref{fig:F1curves}b for YOLO11s-seg, \ref{fig:F1curves}c for YOLO11m-seg, \ref{fig:F1curves}d for YOLO11l-seg, and \ref{fig:F1curves}e displays the curve for YOLO11x-seg.

\subsubsection{\textbf{Evaluation of Image Processing Speeds for YOLO11 Instance Segmentation Algorithm during model training}}
The evaluation of image processing speeds across five YOLO11 configurations reveals significant variations in performance, detailed in Figure \ref{fig:traininginference}a. The YOLO11s-seg model demonstrated the fastest preprocessing time at only 0.2 ms, suggesting an optimized initial data handling capability. Conversely, YOLO11x-seg, despite its higher computational demands, recorded the slowest preprocessing time at 0.4 ms. In terms of inference speed, YOLO11n-seg was the most efficient, processing images in just 3.0 ms, while YOLO11x-seg lagged significantly, requiring 24.8 ms. This indicates a trade-off between depth of model complexity and real-time performance capability. For postprocessing tasks, YOLO11l-seg was the quickest, completing in 1.0 ms, highlighting its streamlined backend processing efficiency.

\subsubsection{\textbf{Evaluation of Convolution Layers, Parameters, Training Time and GFLOPs}}
\begin{figure*}[ht]
\centering
\includegraphics[width=0.90\linewidth]{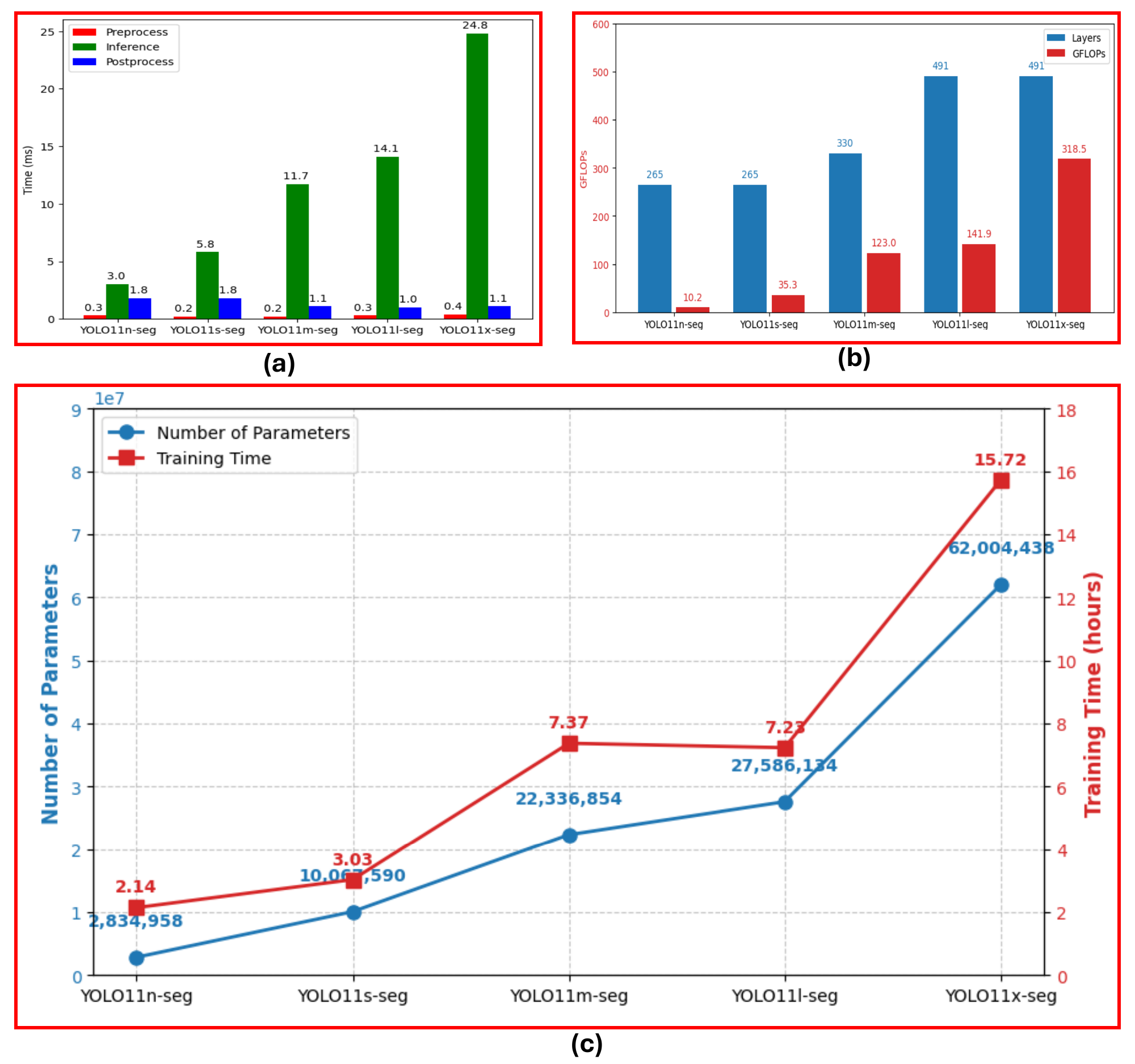}
\caption{a) Bar chart displaying image processing speeds for five YOLO11 configurations, highlighting variations in preprocessing, inference, and postprocessing times. b) Comparison of convolutional layers and GFLOPs across YOLO11 variants, illustrating computational complexity and processing demand. c) Bar chart showing the relationship between training time and the number of parameters for each YOLO11 configuration}
\label{fig:traininginference}
\end{figure*}

The assessment of YOLO11 configurations revealed significant variability in the convolutional layers, parameters, and computational demands across different models, as depicted in  Figures \ref{fig:traininginference}b and \ref{fig:traininginference}c. Both YOLO11l-seg and YOLO11x-seg featured the highest number of layers at 491 (Figure \ref{fig:traininginference}b), indicating a more complex model architecture compared to the rest. In terms of parameters, YOLO11x-seg exhibited the highest count with 62,004,438 parameters (Figure \ref{fig:traininginference}c), substantially more than the other configurations, reflecting its capacity for handling more detailed and intensive processing tasks. Conversely, YOLO11n-seg had the least number of parameters at 2,834,958, suggesting a more streamlined model suited for less computationally intensive applications. When analyzing GFLOPs (Figure \ref{fig:traininginference}b), which serve as an indicator of computational load per image, YOLO11x-seg again recorded the highest at 318.5 GFLOPs, underscoring its substantial processing capability. In stark contrast, YOLO11n-seg maintained the lowest computational burden with only 10.2 GFLOPs, marking it as the most efficient among the evaluated configurations in terms of computational resource usage.

\subsubsection{\textbf{Early Stopping and Model Size}}
During the training phase of the YOLO11 configurations, a maximum of 500 epochs was set, with early stopping criteria applied to prevent overfitting and optimize computational resources. Specifically, YOLO11n-seg ceased training at 485 epochs, YOLO11s-seg at 427 epochs, and YOLO11l-seg at 417 epochs, indicating adequate convergence before reaching the maximum epochs. In contrast, YOLO11m-seg and YOLO11x-seg utilized the full 500 epochs, suggesting a need for extended training to achieve optimal model performance. Following the completion of training, the resultant model sizes were as follows: YOLO11n-seg at 6.1 MB, YOLO11s-seg at 20.6 MB, YOLO11m-seg at 45.2 MB, YOLO11l-seg at 55.9 MB, and YOLO11x-seg at 124.8 MB.

\subsection{\textbf{Validation Results of YOLO11-CBAM model in Dormant Season Dataset}}
\begin{figure*}[ht]
\centering
\includegraphics[width=0.85\linewidth]{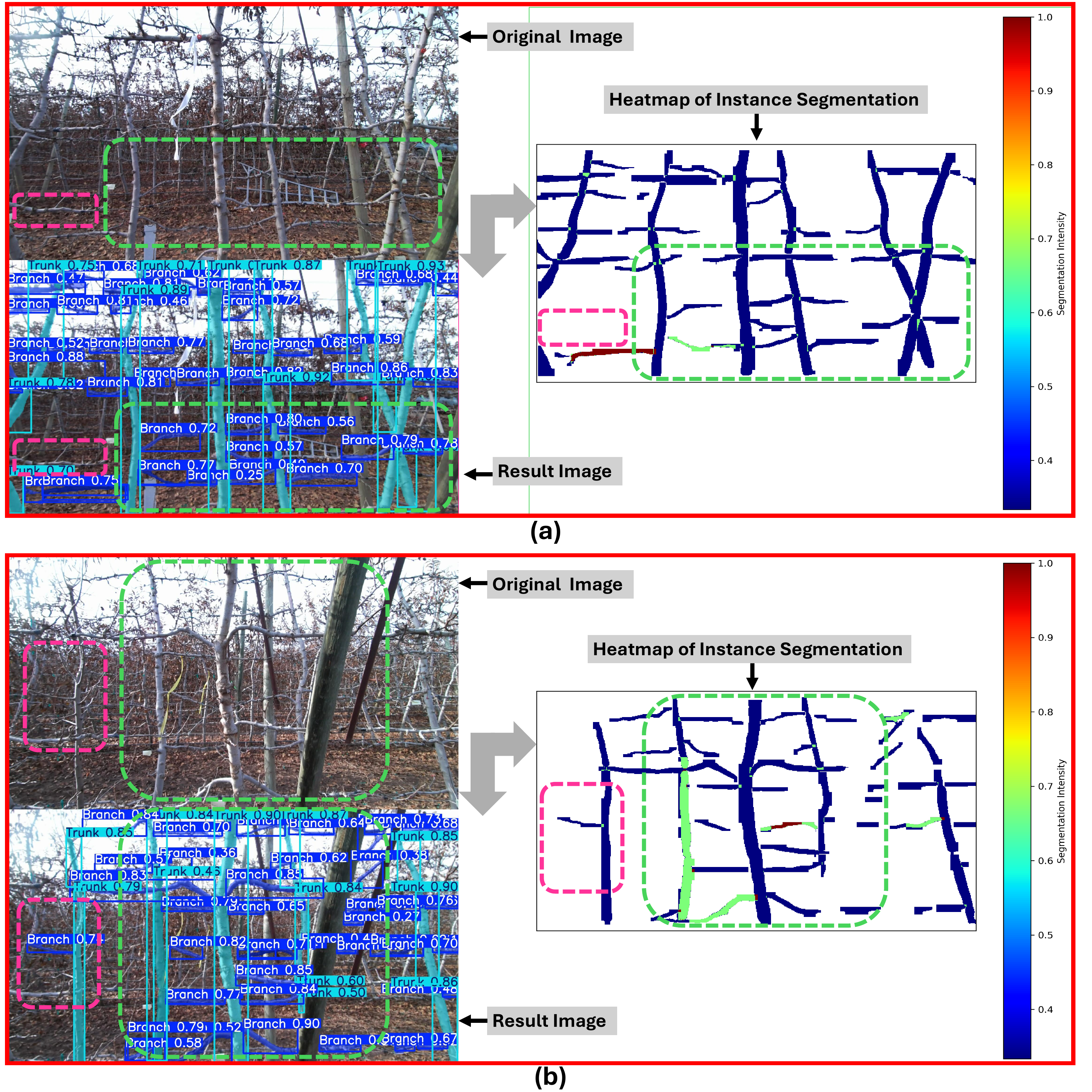}
\caption{Showing the Validation example of YOLO11-CBAM on dormant season dataset: (a) effective trunk and branch segmentation within green, overlooked branch in pink; (b) accurate segmentation amidst similar objects, with missed branches in pink highlighting areas for model improvement. }
\label{fig:resultDormantImage}
\end{figure*}

Figure \ref{fig:resultDormantImage} illustrates the validation results of the YOLO11-CBAM model on a dormant season dataset comprising 78 images. In Figure \ref{fig:resultDormantImage}a, the model demonstrates effective detection and segmentation capabilities, accurately identifying the majority of trunk and branch areas within the green dotted rectangle. However, within the pink dotted rectangle, the model overlooks a branch segment prominently positioned in the foreground. This figure includes an original RGB image from a Microsoft Azure Kinect vision camera, the segmentation result below it, and a heatmap representation of the extracted trunk and branch masks to the right. Similarly, Figure \ref{fig:resultDormantImage}b provides another instance of the model's performance during dormant season validation. Here, the green dotted rectangle encapsulates areas where the model accurately segmented trunks and branches, distinguishing between actual tree components and an orchard training pole, which it correctly did not classify as a trunk due to its dissimilar physical attributes. Conversely, in the pink rectangle, the model, while successfully recognizing a trunk, fails to identify three additional branches present in the vicinity. These instances indicate potential areas for improvement in model training. Enhancing the training dataset, particularly by augmenting the number of samples, currently at 859 images spanning both dormant and canopy seasons could significantly reduce such detection inaccuracies.

\textbf{Performance Metrics Validation:} During the validation of completely dormant season images, the YOLO11-CBAM model demonstrated significant performance variations across different classes. For the "All" class, YOLO11x-seg recorded the highest mask precision at 0.91. YOLO11m-seg exhibited superior performance in terms of mask recall and mAP@50, scoring 0.84 and 0.88, respectively. In the "Branch" class, YOLO11x-seg again led in mask precision with a score of 0.87, while YOLO11m-seg topped mask recall at 0.75 and mAP@50 at 0.81. For the "Trunk" class, both YOLO11m-seg and YOLO11x-seg tied for the highest mask precision at 0.94 and mask recall at 0.92, with YOLO11m-seg also achieving the highest mAP@50 of 0.96. These metrics highlight the nuanced capabilities of each configuration in handling specific challenges posed by dormant season imagery. Comprehensive details of box and mask metrics across all five YOLO11 configurations integrated with the CBAM are presented in Table \ref{tab:validation_metrics_dormant}, underscoring the varied strengths of each model configuration in this specialized validation context.

\begin{table*}[ht]
\centering
\caption{\textbf{Validation Metrics for Dormant Season Across YOLO11-CBAM Configurations}}
\label{tab:validation_metrics_dormant}
\begin{tabular}{@{}llccccccccc@{}}
\toprule
\textbf{Model} & \textbf{Class} & \multicolumn{3}{c}{\textbf{Box Metrics}} & \multicolumn{3}{c}{\textbf{Mask Metrics}} \\
 & & \textbf{Precision} & \textbf{Recall} & \textbf{mAP@50} & \textbf{Precision} & \textbf{Recall} & \textbf{mAP@50} \\
\midrule
\multirow{3}{*}{YOLO11n-seg} 
& All & 0.877 & 0.817 & 0.874 & 0.844 & 0.777 & 0.827 \\
& Branch & 0.856 & 0.731 & 0.814 & 0.795 & 0.664 & 0.730 \\
& Trunk & 0.897 & 0.903 & 0.933 & 0.894 & 0.891 & 0.924 \\
\midrule
\multirow{3}{*}{YOLO11s-seg} 
& All & 0.895 & 0.848 & 0.903 & 0.876 & 0.792 & 0.847 \\
& Branch & 0.880 & 0.785 & 0.858 & 0.831 & 0.696 & 0.761 \\
& Trunk & 0.911 & 0.910 & 0.947 & 0.921 & 0.887 & 0.934 \\
\midrule
\multirow{3}{*}{YOLO11m-seg} 
& All & 0.940 & 0.898 & 0.935 & 0.903 & 0.842 & 0.886 \\
& Branch & 0.922 & 0.846 & 0.901 & 0.859 & 0.757 & 0.813 \\
& Trunk & 0.958 & 0.949 & 0.969 & 0.948 & 0.927 & 0.960 \\
\midrule
\multirow{3}{*}{YOLO11l-seg} 
& All & 0.906 & 0.730 & 0.794 & 0.875 & 0.695 & 0.756 \\
& Branch & 0.891 & 0.677 & 0.755 & 0.841 & 0.622 & 0.685 \\
& Trunk & 0.921 & 0.783 & 0.834 & 0.908 & 0.767 & 0.827 \\
\midrule
\multirow{3}{*}{YOLO11x-seg} 
& All & 0.937 & 0.887 & 0.934 & 0.910 & 0.832 & 0.884 \\
& Branch & 0.915 & 0.828 & 0.897 & 0.871 & 0.742 & 0.808 \\
& Trunk & 0.960 & 0.947 & 0.972 & 0.949 & 0.923 & 0.959 \\
\bottomrule
\end{tabular}
\end{table*}

\begin{figure}[ht]
\centering
\includegraphics[width=0.85\linewidth]{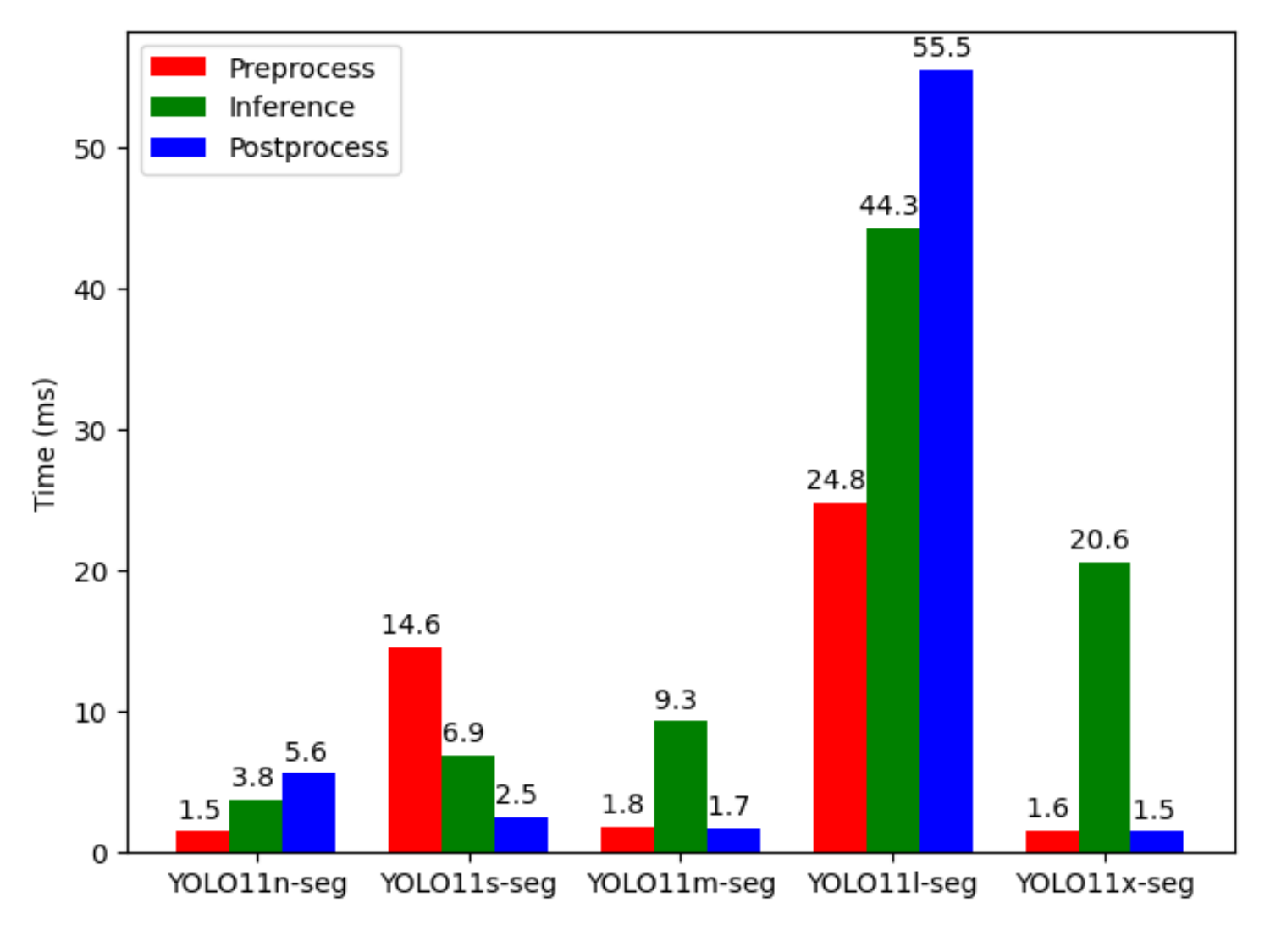}
\caption{ Image processing speeds for the dormant season validation of tree trunk and branch segmentation using the five YOLO11-CBAM Instance Segmentation configurations}
\label{fig:dormantImagespeeds}
\end{figure}
\textbf{Image Processing Speeds:}During the validation of dormant season images, distinct disparities in image processing speeds were observed among the YOLO11-CBAM configurations. The YOLO11l-seg configuration demonstrated the slowest processing speeds across all stages: 24.8 ms in preprocessing, 44.3 ms in inference, and 55.5 ms in postprocessing, indicating a significant computational demand. In contrast, the YOLO11x-seg configuration exhibited the fastest postprocessing speed at only 1.5 ms, while YOLO11n-seg managed the swiftest inference time at 3.8 ms. These variations highlight the trade-offs between processing speed and computational load in different model configurations. For a detailed visualization of these processing speeds across the configurations, refer to Figure \ref{fig:dormantImagespeeds}, which illustrates the efficiency and delays inherent in each model during the validation on 78 dormant season images.

\subsection{\textbf{Validation Results of Canopy Season Dataset}}
\begin{figure*}[ht]
\centering
\includegraphics[width=0.85\linewidth]{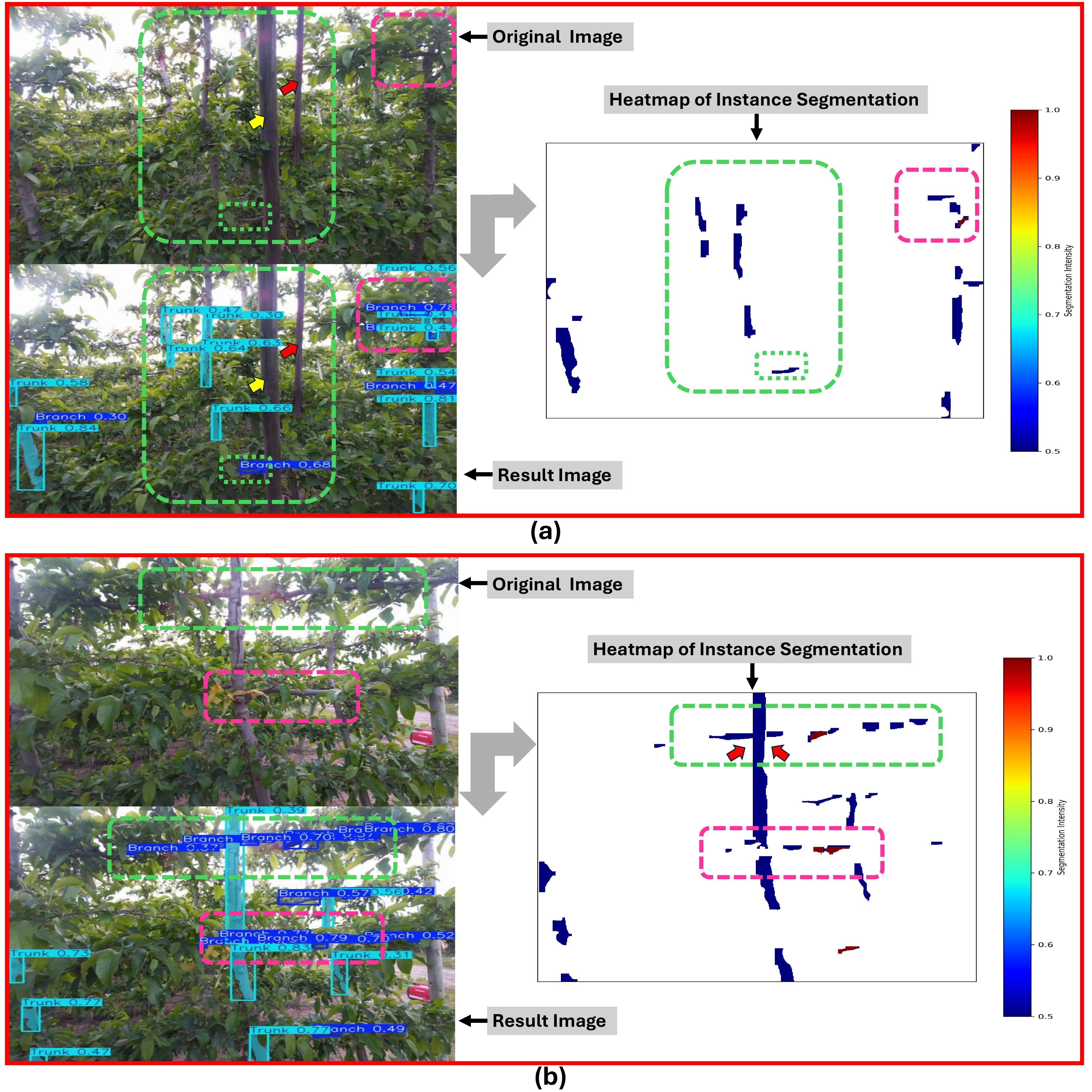}
\caption{Segmentation performance of the YOLO11-CBAM during key seasonal operations, highlighting areas of weakness (pink) and strength (green): (a) Pre-blossom season: effective trunk segmentation vital for collision-free dormant pruning. (b) Flower blossom season: precise trunk and branch identification crucial for automated thinning and robotic pollination  }
\label{fig:resultCanopyImage}
\end{figure*}
Figure \ref{fig:resultCanopyImage} illustrates the capabilities of the YOLO11-CBAM model in segmenting tree structures during the canopy season, despite the challenging visibility conditions caused by dense foliage. During this period, the canopy is at its most voluminous, filled with thick, green leaves before transitioning to a less dense state with yellowing leaves. The segmentation effectiveness is demonstrated in Figure \ref{fig:resultCanopyImage}a, where the upper left image represents the original scene captured by the imaging system, and the lower left image displays the segmentation results achieved by the YOLO11n configuration. In these results, the model proficiently identified tree trunks, a critical task given the complexity of the scene. Notably, the model was able to differentiate between actual tree trunks and similar-looking objects such as training poles and metal supports for training wires, which are indicated by yellow and red arrows respectively in the larger green dotted rectangle. This region highlights the model's precision in distinguishing true trunks from other elements with similar color profiles in the RGB image space. Furthermore, the smaller dotted rectangle and the pink dotted area reveal that the model successfully extracted some branch-level details, underscoring the YOLO11-CBAM fusion's effectiveness in handling the intricate details necessary for precise agricultural segmentation during peak foliage.

\par
Figure \ref{fig:resultCanopyImage}b presents further evidence of the YOLO11-CBAM model's adeptness in segmenting trunks and branches during the canopy season. This figure displays the segmentation accuracy where the model not only identifies the trunks prominently positioned in the foreground but also discerns branch areas. The original image on the upper left, captured using a consumer-level camera within a commercial scilate orchard, contrasts with the lower left image that illustrates the segmentation outcomes. The right side of the figure displays the extracted RGB heatmap, highlighting the precision of trunk and branch detection despite the complex background of dense foliage. In the green and pink dotted rectangles, the model showcases its capability to detect critical branch areas, which are vital for automated crop load management decisions. While the model successfully captures the significant regions necessary for robotic operations, minor segmentation challenges are evident, such as the presence of yellow research tape, which introduces slight inaccuracies. These segmented outputs are crucial for developing advanced applications like collision-free robotic motion planning. Furthermore, the detailed branch data extracted can be utilized to automate decisions like green fruit thinning by leveraging limb cross-sectional area (LCSA) calculations, a technique previously explored in Ahmed et al. (2023) \cite{ahmed2023machine}. 

\textbf{Performance Metrics Validation:}
During the validation of the canopy season dataset, segmentation performance metrics were meticulously analyzed for the "All," "Branch," and "Trunk" classes. In the "All" class, YOLO11s-seg displayed superior precision, achieving a score of 0.579. This configuration also led in the "Branch" class, registering the highest precision at 0.516, and continued its dominance in the "Trunk" class with a precision of 0.643. Regarding recall, YOLO11n-seg and YOLO11s-seg jointly recorded the highest for the "All" class at 0.416. In contrast, YOLO11x-seg excelled in the "Branch" class with a recall of 0.52, while YOLO11n-seg led the "Trunk" class with a recall of 0.50. The model's ability to precisely map the area under the curve for the intersection over union at the 50\% threshold (mAP@50) was also noteworthy. YOLO11s-seg outperformed in the "All" class with a mAP@50 of 0.43, maintained the lead in the "Branch" class with 0.34, and scored highest in the "Trunk" class at 0.52. These results demonstrate the effectiveness of the YOLO11 configurations when integrated with the CBAM, especially in complex canopy environments where accurate segmentation is critical. Detailed evaluations of both box and mask metrics across all five YOLO11 configurations are thoroughly documented in Table \ref{tab:canopy_validation_metrics}, providing a comprehensive overview of the model's performance across different tree categories during the canopy season.
\begin{table*}[ht]
\centering
\caption{\textbf{Validation Metrics on Canopy Season Images for YOLO11-CBAM Configurations}}
\label{tab:canopy_validation_metrics}
\begin{tabular}{@{}lcccccccccc@{}}
\toprule
\textbf{Configuration} & \textbf{Class} & \multicolumn{3}{c}{\textbf{Box Metrics (P, R, mAP@50)}} & \multicolumn{3}{c}{\textbf{Mask Metrics (P, R, mAP@50)}} \\
\cmidrule(lr){3-5} \cmidrule(lr){6-8}
 & & \textbf{Precision} & \textbf{Recall} & \textbf{mAP@50} & \textbf{Precision} & \textbf{Recall} & \textbf{mAP@50} \\
\midrule
\multirow{3}{*}{YOLO11n-seg} & All & 0.541 & 0.459 & 0.441 & 0.484 & 0.416 & 0.382 \\
 & Branch & 0.517 & 0.399 & 0.369 & 0.422 & 0.326 & 0.284 \\
 & Trunk & 0.565 & 0.520 & 0.513 & 0.546 & 0.505 & 0.479 \\ \hline 
\multirow{3}{*}{YOLO11s-seg} & All & 0.671 & 0.476 & 0.524 & 0.579 & 0.416 & 0.435 \\
 & Branch & 0.645 & 0.421 & 0.466 & 0.516 & 0.342 & 0.345 \\
 & Trunk & 0.697 & 0.531 & 0.582 & 0.643 & 0.490 & 0.525 \\ \hline 
\multirow{3}{*}{YOLO11m-seg} & All & 0.540 & 0.332 & 0.366 & 0.501 & 0.292 & 0.319 \\
 & Branch & 0.538 & 0.285 & 0.337 & 0.448 & 0.220 & 0.253 \\
 & Trunk & 0.542 & 0.379 & 0.395 & 0.554 & 0.365 & 0.386 \\ \hline 
\multirow{3}{*}{YOLO11l-seg} & All & 0.625 & 0.415 & 0.468 & 0.546 & 0.358 & 0.385 \\
 & Branch & 0.635 & 0.370 & 0.444 & 0.503 & 0.287 & 0.317 \\
 & Trunk & 0.615 & 0.460 & 0.491 & 0.589 & 0.429 & 0.453 \\ \hline 
\multirow{3}{*}{YOLO11x-seg} & All & 0.575 & 0.475 & 0.510 & 0.569 & 0.378 & 0.415 \\
 & Branch & 0.1066 & 0.550 & 0.422 & 0.199 & 0.523 & 0.302 \\
 & Trunk & 0.601 & 0.529 & 0.559 & 0.615 & 0.454 & 0.501 \\
\bottomrule
\end{tabular}
\end{table*}

\begin{figure}[ht]
\centering
\includegraphics[width=0.75\linewidth]{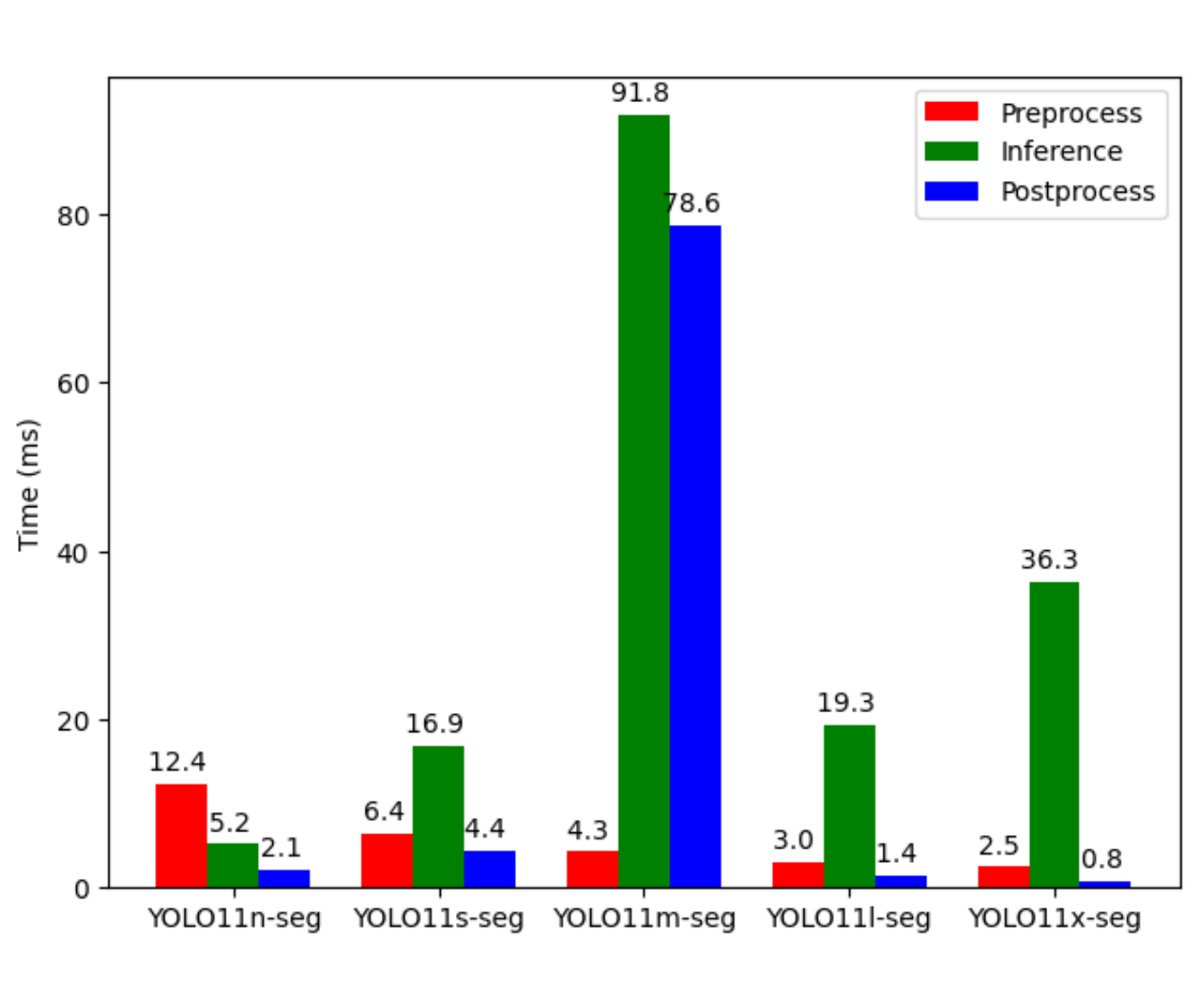}
\caption{Bar diagram showing Image processing speeds for YOLO11-CBAM configurations during canopy season validation}
\label{fig:ImgSpeedResultCanopyValidation}
\end{figure}

\textbf{Image Processing Speeds:}
In the validation of canopy season images using YOLO11 configurations, the image processing speeds varied significantly across different stages. YOLO11x-seg exhibited the fastest preprocessing time at 2.5 ms, while YOLO11m-seg recorded the slowest inference time, taking 91.8 ms. In postprocessing, YOLO11x-seg was again the most efficient, completing the task in just 0.8 ms, in contrast to YOLO11m-seg, which took the longest at 78.6 ms. These results underline the variability in processing efficiency among the different configurations of the YOLO11-CBAM fusion, showcasing both highly efficient and slower processing phases. For a detailed visual comparison of image processing speeds across all configurations, refer to Figure \ref{fig:ImgSpeedResultCanopyValidation}.  

\section{Discussion}
\begin{figure}[ht]
\centering
\includegraphics[width=0.95\linewidth]{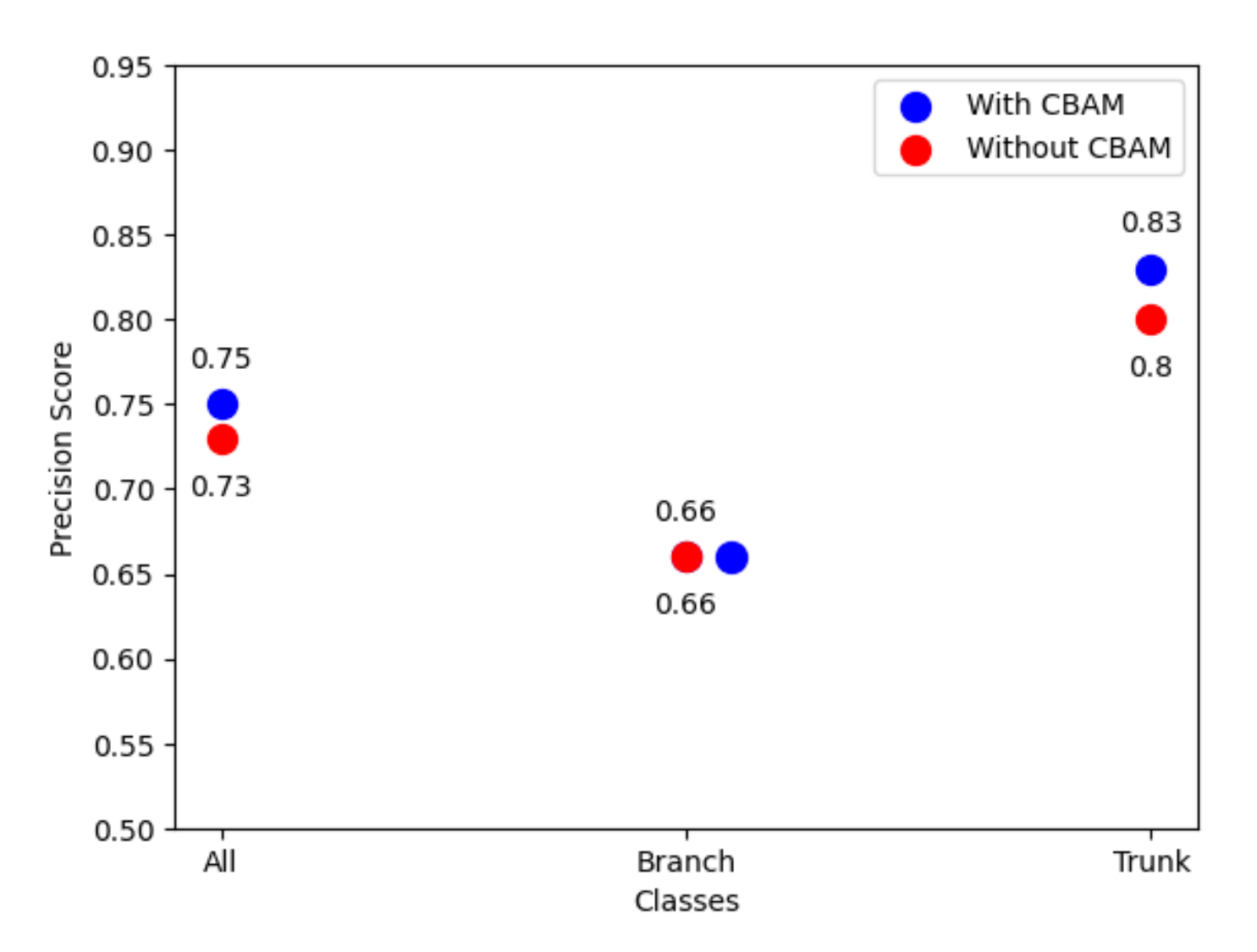}
\caption{Comparing precision scores of YOLO11m-seg with and without CBAM for trunk, branch, and all classes}
\label{fig:PrecisionYOLO11m}
\end{figure}
\begin{figure*}[ht]
\centering
\includegraphics[width=0.95\linewidth]{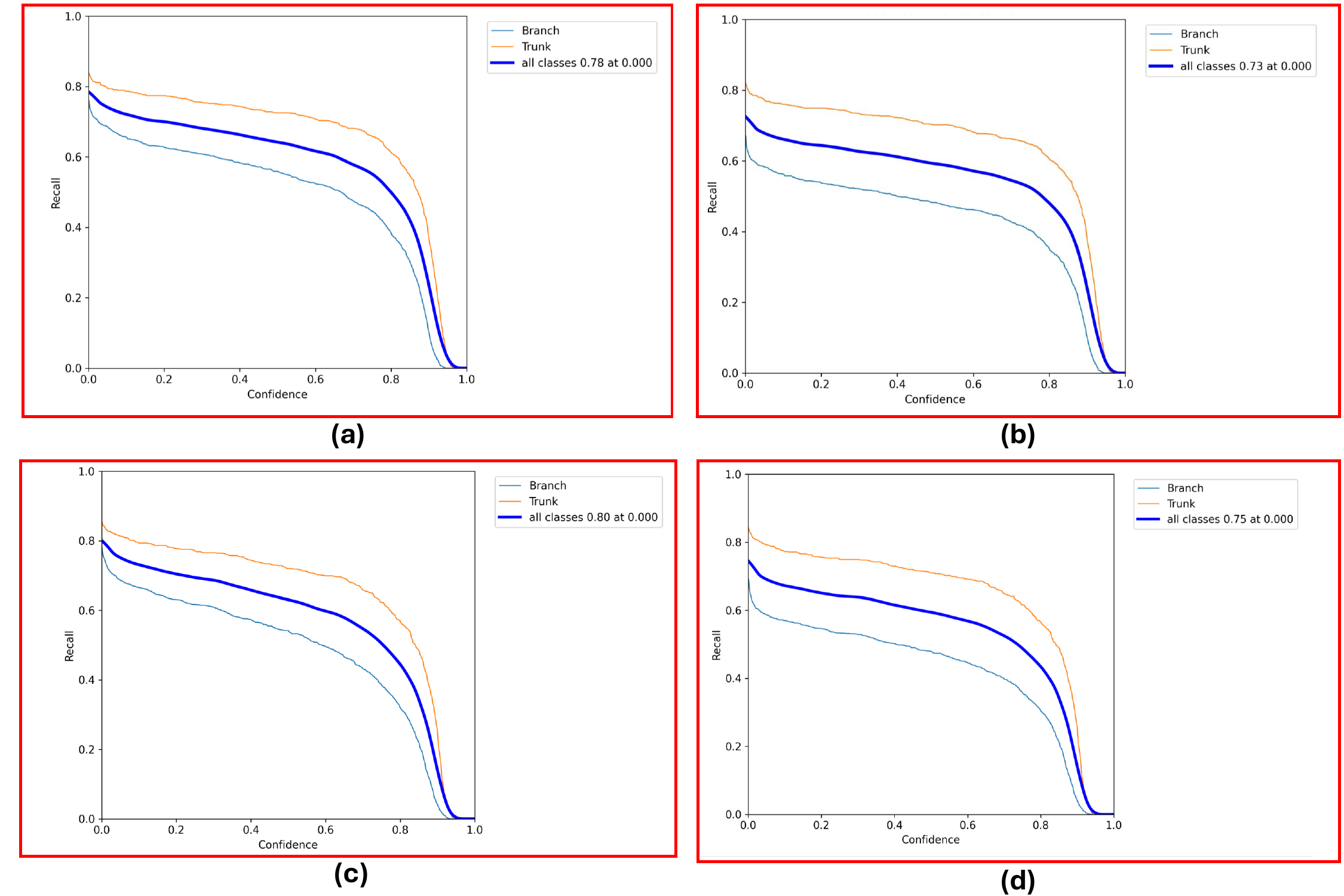}
\caption{Recall Confidence Curves for the YOLO11x-seg configuration, illustrating the highest recall scores achieved with and without CBAM integration across different classes: (a) Box Recall Confidence Curve for YOLO11x-seg without CBAM; (b) Mask Recall Confidence Curve for YOLO11x-seg without CBAM; (c) Box Recall Confidence Curve for YOLO11x-seg with CBAM integration;  (d) Mask Recall Confidence Curve for YOLO11x-seg with CBAM integration}
\label{fig:Recalls}
\end{figure*}
The highest precision score during training was recorded by the YOLO11m-seg model configuration, particularly notable in the trunk class with a score of 0.83. This superior performance underscores the model's adeptness at trunk segmentation, a critical capability for precision agricultural applications. 
A comprehensive comparison of YOLO11m-seg's performance with and without the Convolutional Block Attention Module (CBAM) is detailed in Figure \ref{fig:PrecisionYOLO11m}. This figure visually demonstrates the enhancements CBAM brings to precision, highlighting its value in improving model accuracy for specific classes.

As depicted in Figure \ref{fig:Recalls}, the recall confidence curves highlight the superior recall values achieved by the YOLO11xseg configuration when trained with CBAM compared to its counterparts without CBAM integration. Specifically, in the ”All” class, the YOLO11x-seg achieved a box recall of 0.66 without CBAM, which slightly increased to 0.67 with CBAM. This enhancement was also observed in the ”Branch” classe, where box recalls improved from 0.57 to 0.59, indicating consistent performance improvements across different structural elements of the trees. Similarly, mask recall metrics, which provide insights into the segmentation accuracy, demonstrated noticeable improvements with CBAM integration. For the ”All” class, mask recall increased from 0.60 to 0.61; in the ”Branch” class, it rose from 0.49 to 0.50; and in the ”Trunk” class, it advanced from 0.71 to 0.72 with CBAM. 

\begin{figure*}[tp]
\centering
\includegraphics[width=0.58\linewidth]{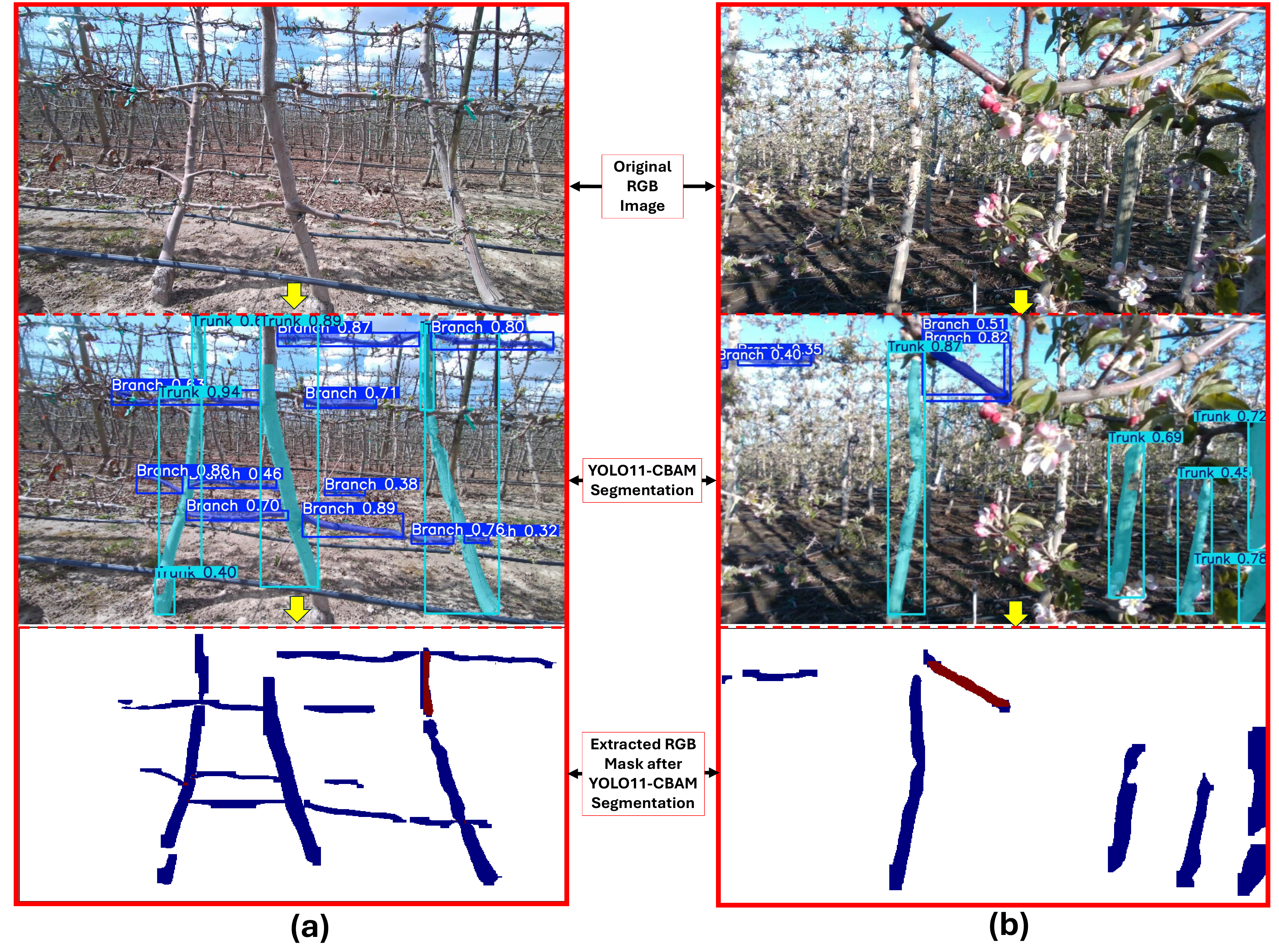}
\caption{Segmentation performance of the YOLO11 model during key seasonal operations. (a) Pre-blossom season showing effective trunk segmentation, vital for collision-free dormant pruning. (b) Flower blossom season illustrating precise identification of trunks and branches, crucial for automated thinning and collision-free robotic pollination.}
\label{fig:discussion1}
\end{figure*}

\begin{figure*}[ht]
\centering
\includegraphics[width=0.58\linewidth]{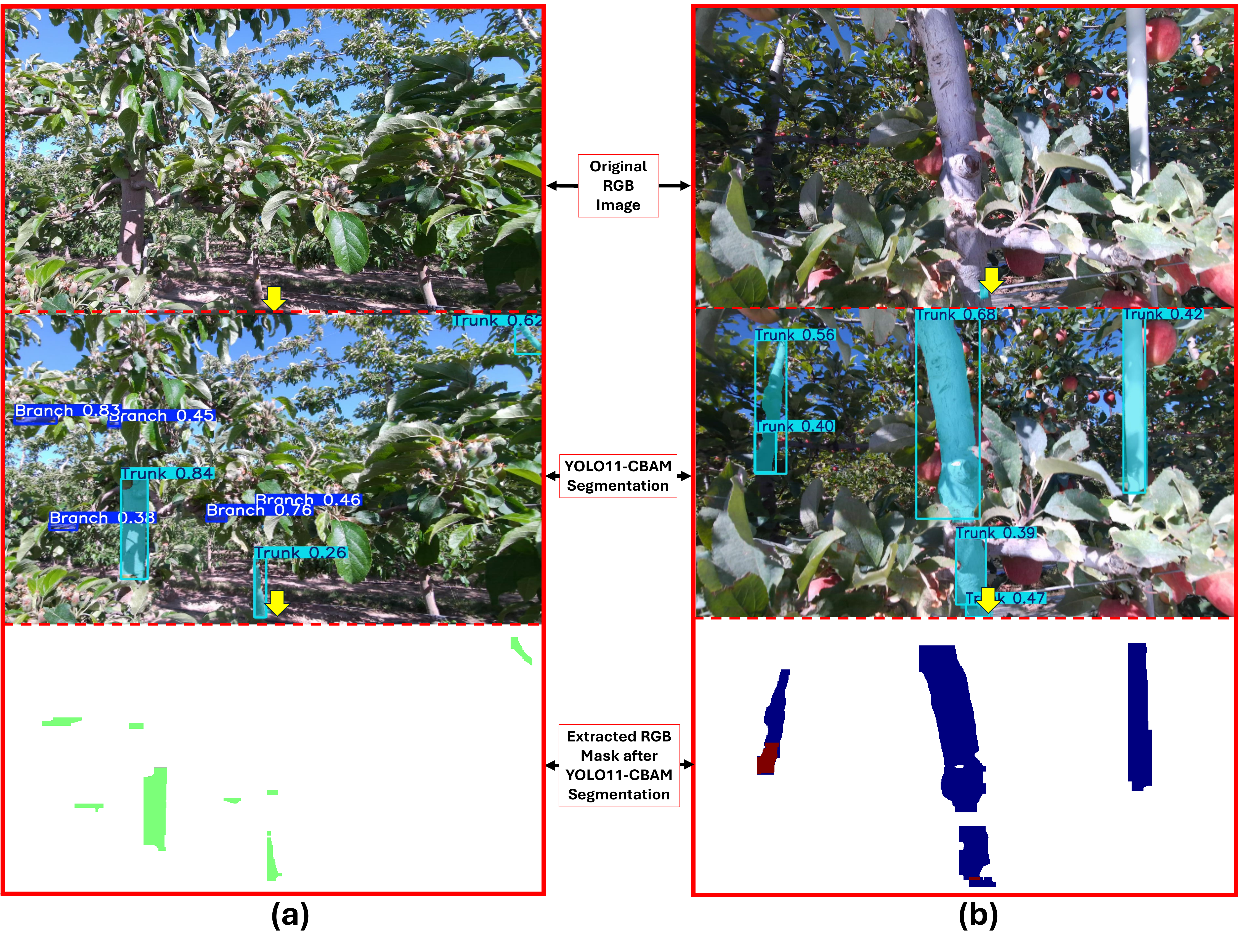}
\caption{Segmentation performance of the YOLO11 model during key seasonal operations. (a) Immature green fruit thinning season, during the month of June. (b) Harvest Season, during the month of October}
\label{fig:discussion2}
\end{figure*}

This model was further subjected to testing across four distinct seasonal variations not previously encountered during training. This included the pre-flower bloom in February 2023, characterized by nascent leaf development as depicted in Figure \ref{fig:discussion1}a; the peak bloom period in April 2023, illustrated in Figure \ref{fig:discussion1}b; the critical green fruit thinning phase in June 2024, optimal for decision-making at diameters under 25 mm, shown in Figure \ref{fig:discussion2}a; and the harvest season in October 2024, captured in Figure \ref{fig:discussion2}b.

This model was further subjected to testing across four distinct seasonal variations not previously encountered during training. This included the pre-flower bloom in February 2023, characterized by nascent leaf development as depicted in Figure \ref{fig:discussion1}a; the peak bloom period in April 2023, illustrated in Figure \ref{fig:discussion1}b; the critical green fruit thinning phase in June 2024, optimal for decision-making at diameters under 25 mm, shown in Figure \ref{fig:discussion2}a; and the harvest season in October 2024, captured in Figure \ref{fig:discussion2}b.  
\subsection{Pre-Blossom Season}
The segmentation performance of the YOLO11 model during the pre-blossom season, as shown in Figure \ref{fig:discussion1}a, reveals its effectiveness in identifying tree trunks, which are more visible compared to branches. This visibility significantly aids the model's capability to segment trunks accurately, an essential feature for dormant pruning operations. While some branches are also identified in these results, their recognition is crucial for the development of collision-free robotic systems in orchards such as earlier studies reports \cite{you2023semiautonomous, he2018sensing, zahid2020development }. Such automated pruning systems require precise navigation capabilities to avoid damaging tree structures during field operation, making the accurate detection and segmentation of both trunks and branches vital. The model's ability to distinguish these components in the orchard underlines its potential utility in enhancing robotic operations, specifically in ensuring that pruning robots can perform their tasks without interference from the physical environment.

\subsection{During Flower Blossom Season}
During the flower blossom season, the results depicted in Figure \ref{fig:discussion1}b demonstrate the model's continued effectiveness in segmenting trunks and branches amidst the complexity of blooming flowers. This capability is particularly significant for flower thinning robots, which operate in this critical period to ensure optimal fruit production. The precision in segmentation allows these robots to perform thinning tasks without damaging the blossoms, contributing to efficient and effective crop management. Additionally, the accurate detection of tree components during this season is crucial for collision-free robotic operations in precision pollination. With the increasing focus on robotic pollination, as highlighted in studies like \cite{sapkota2023robotic, sapkota2024robotics}, ensuring that robots can navigate and operate without collision during this delicate phase is essential. This not only enhances pollination effectiveness but also minimizes potential damage to the trees and blossoms, thereby supporting advanced agricultural practices through the integration of robust image processing technologies.

\subsection{Green Fruit Thinning Season}
During the green fruit thinning season, as demonstrated in Figure \ref{fig:discussion2}a, the YOLO11 model showcases its capability to identify trunks and branches within heavily dense canopy cover. Despite the complexity of the foliage, the model successfully segments critical structural elements of the tree, which are essential for robotic navigation and operational decision-making in orchards. This ability is particularly crucial for green fruit thinning operations, where decisions on fruit load are often based on the branch diameter (LCSA). Accurate detection and segmentation of tree branches are vital to determining branch diameter, which directly influences thinning decisions to maintain optimal fruit counts per LCSA, as discussed in studies like \cite{chen2018multi, brown2024tree}. 

\subsection{Harvest Season}
In the apple harvest season, depicted in Figure \ref{fig:discussion2}b, effective trunk segmentation by the YOLO11 model plays a pivotal role in enabling robotic harvesters to navigate through the orchard without colliding with the trees. This feature is particularly significant as it ensures the safety and efficiency of the harvesting process. Although some branches in the forefront of the image were not identified, potentially due to bright lighting conditions and obstructive foliage, these limitations can be addressed by augmenting the training dataset. Increasing the diversity and quantity of training samples could improve the model’s ability to recognize branches under varying environmental conditions.

Our methodology distinctly stands out by not only focusing on trunk detection accuracy but also testing the robustness of our integrated YOLO11-CBAM model across multiple seasonal conditions. This comprehensive approach ensures our model's adaptability and performance consistency in real-world agricultural settings, making it uniquely robust compared to the single-season or condition-specific focus seen in previous studies. Recent studies have developed various methodologies for tree trunk detection, each contributing unique insights into enhancing robotic operations in diverse environments. For instance, \cite{da2022edge} focused on benchmarking multiple deep learning models for forestry monitoring, demonstrating the strength of YOLOR in achieving high F1 scores. \cite{juman2016novel} innovated a novel detection method combining color imaging and depth information for oil-palm plantation navigation, significantly reducing false positives. In structured orchards, \cite{bargoti2015pipeline} used a combination of lidar and image data to map apple orchards efficiently, while \cite{lamprecht2015atrunk} employed an ALS-based approach to isolate trunk points accurately. Lastly, \cite{li2023tree} introduced a multiscale attention-based deep learning method that excels in urban tree trunk segmentation, handling diversity and intricate structures effectively.

\section{Conclusion and Future}
In this study, we evaluated the capability the YOLO11 model integrated with the CBAM across multiple seasonal conditions within commercial apple orchards. The objective was to not merely assess the model's performance in isolation but to validate its effectiveness through various stages of the agricultural cycle, including dormant, pre-blossom, flower blossom, green fruit thinning, and harvest seasons. This approach allowed us to demonstrate the model's capability in handling complex agricultural segmentation tasks under diverse environmental conditions. Utilizing a mixed dataset that included images from both dormant and canopy seasons, the YOLO11-CBAM model showcased significant improvements in detection and segmentation accuracy, particularly in identifying tree trunks and branches critical for robotic agricultural operations. The highest recall and precision metrics were observed in the YOLO11x-segCBAM and YOLO11m-segCBAM  s respectively, emphasizing its efficiency and potential for real-time applications. The validation results affirmed the model’s proficiency in precise segmentation, essential for enhancing the operational efficacy of robotic systems in orchards. For instance, during the canopy season, the model effectively navigated through dense foliage, proving its indispensable role in tasks such as green fruit thinning and harvest navigation, where accurate localization of structural tree elements is paramount. The specific conclusions of this experiment are:
\begin{itemize}
    \item \textbf{Training Performance:} YOLO11x-seg achieved the highest recall values during the mixed dataset training phase, with box recall scores of 0.67 (All class), 0.59 (Branch class), and 0.75 (Trunk class). Similarly, mask recall scores were 0.61 (All class), 0.50 (Branch class), and 0.72 (Trunk class), showcasing its effectiveness in diverse seasonal conditions.
    \item \textbf{Seasonal Adaptability: }The model demonstrated strong performance across both dormant and canopy seasons, successfully identifying and segmenting tree trunks and branches. This adaptability is crucial for year-round agricultural robotic operations.
    \item \textbf{Dormant Season Performance:} In the dormant season validation, YOLO11x-seg exhibited notable precision, especially in the Trunk class, where it achieved a mask precision of 0.94 and a mask recall of 0.92, facilitating reliable dormant pruning applications.
    \item \textbf{Canopy Season Efficacy: } During the canopy season, the model proficiently handled dense foliage, a challenging scenario for segmentation. YOLO11x-seg maintained a mask precision of 0.615 (Trunk class) and a recall of 0.523 (Branch class), proving its utility in precision agricultural tasks like green fruit thinning.
\end{itemize}
 
As a future directive, enhancing the robustness of the YOLO11-CBAM model for complex orchard environments necessitates a substantial expansion of the training dataset. The initial phase of this research utilized a relatively modest dataset comprising 859 images across various seasonal conditions. To elevate the model's capabilities to an industrial level, a significantly larger and more diverse dataset should be curated and annotated. This expansion will enable the model to learn from a broader array of visual inputs, thereby improving its accuracy and reliability in real-world applications. Moreover, addressing the visibility constraints of branches during the dense canopy season is crucial for year-round operational efficacy. One promising approach is to implement precise image registration techniques that stitch or map dormant season images to their canopy season counterparts for each tree. This method would allow for the consistent visibility of critical tree structures such as branches and trunks throughout different seasons. By ensuring that these key elements are visible and accurately segmented all year round, the model can provide more reliable data for automated agricultural processes. The integration of advanced image processing methodologies with machine learning algorithms presents a significant opportunity to enhance the precision of agricultural robots. This approach not only supports the development of more autonomous and efficient robotic systems but also contributes to the broader goal of sustainable and technologically advanced agriculture. Future research should focus on these areas to ensure that the progress in agricultural robotics keeps pace with the increasing demands of modern orchard management.

\section{Acknowledgment and Funding}
This research is funded by the National Science Foundation and United States Department of Agriculture, National Institute of Food and Agriculture through the “AI Institute for Agriculture” Program (Award No.AWD003473). We extend our heartfelt gratitude to Zhichao Meng, Martin Churuvija and Rounak Pokharel. Special thanks to Dave Allan for granting orchard access.

\textbf{Authors' Contribution:} 
Ranjan Sapkota : conceptualization, data curation, software, methodology, validation,  writing original draft.  Manoj Karkee (correspondence)  review, editing and overall funding to supervisory 
\bibliography{references.bib}
\bibliographystyle{ieeetr}
\begin{IEEEbiography}
[{\includegraphics[width=1in,height=0.95in,clip,keepaspectratio]{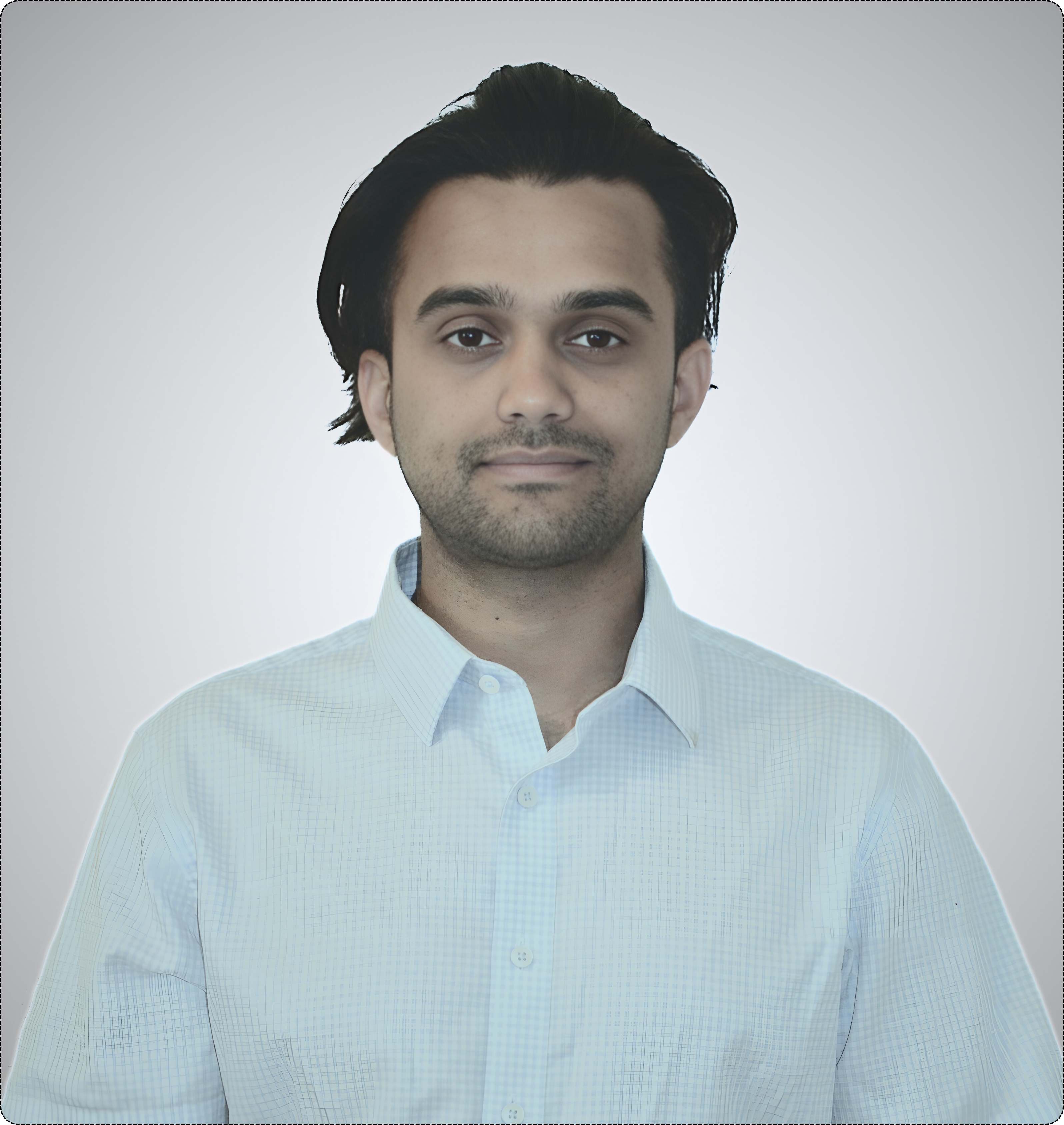}}]{Ranjan Sapkota} Mr. Sapkota is a Ph.D. student at Washington State University in the Department of Biological Systems Engineering and Center for Precision and Automated Agricultural Systems. His research primarily focuses on Automation and Robotics for Agriculture, utilizing Artificial Intelligence, Large Language Models, Machine Vision, Robot Manipulation Systems, Deep Learning, Machine Learning and Generative AI Technologies.  He obtained his B.Tech degree in Electrical and Electronics Engineering from Uttarakhand Technical University, India in 2019. He then pursued his MS in Agricultural and Biosystems Engineering from North Dakota State University from 2020 to 2022.
\end{IEEEbiography}
\begin{IEEEbiography}
[{\includegraphics[width=1in,height=0.95in,clip]{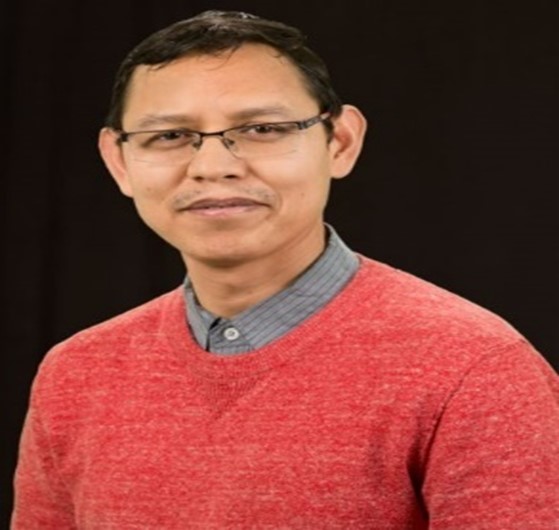}}]{Prof. Dr. Manoj Karkee} Dr. Karkee currently serves as the Professor and Director of the Center for Precision and Automated Agricultural Systems at Washington State University. His research focuses on agricultural automation and mechanization programs, with an emphasis on machine vision-based sensing, automation and robotic technologies for specialty crop production. He obtained his BS in Computer Engineering from Tribhuwan University in 2002. He pursued his MS in Remote Sensing and Geographic Information Systems at Asian Institute of Technology, Thailand, and earned his Doctorate in Agricultural Engineering and Human-Computer Interaction from Iowa State University in 2009. 
\end{IEEEbiography}
\end{document}